\documentclass{article}

\usepackage{microtype}
\usepackage{graphicx}
\usepackage{subfigure}
\usepackage{booktabs} % for professional tables

\usepackage{hyperref}

\usepackage[accepted]{icml2024}

\usepackage{amsmath}
\usepackage{amssymb}
\usepackage{mathtools}
\usepackage{amsthm}

\usepackage[capitalize,noabbrev]{cleveref}

\theoremstyle{plain}

\newtheorem{proposition}{Proposition}

\theoremstyle{definition}

\newtheorem{assumption}{Assumption}
\theoremstyle{remark}

\usepackage[textsize=tiny]{todonotes}

\usepackage{bm}
\usepackage{booktabs}
\usepackage{listings}
\usepackage{extra_styles}
\usepackage{multirow}
\usepackage{array}
\usepackage{paralist}
\usepackage{comment}        %
\usepackage{bbm}
\usepackage[normalem]{ulem}
\usepackage[accsupp]{axessibility}  %
\usepackage{algorithm}
\usepackage{algorithmic}

\crefname{section}{Sec.}{Secs.}
\Crefname{section}{Section}{Sections}
\Crefname{table}{Table}{Tables}
\crefname{table}{Tab.}{Tabs.}
\Crefname{figure}{Figure}{Figures}
\crefname{figure}{Fig.}{Figs.}
\Crefname{equation}{Equation}{Equations}
\crefname{equation}{Eq.}{Eqs.}
\Crefname{algorithm}{Algorithm}{Algorithms}
\crefname{algorithm}{Alg.}{Algs.}
\Crefname{proposition}{Proposition}{Propositions}
\crefname{proposition}{Prop.}{Props.}
\Crefname{appendix}{Appendix}{Appendices}
\crefname{appendix}{App.}{Apps.}

\usepackage{xcolor}
\usepackage{colortbl}
\definecolor{mygray}{gray}{.9}

{ \bgroup
  \addtolength\abovedisplayshortskip{#1}
  \addtolength\abovedisplayskip{#1}
  \addtolength\belowdisplayshortskip{#1}
  \addtolength\belowdisplayskip{#1}}
{\egroup\ignorespacesafterend}

\newcommand{\dif}{\mathop{}\!\mathrm{d}}

\newcommand{\name}{SimPro\xspace}

\icmltitlerunning{SimPro: A Simple Probabilistic Framework Towards Realistic Long-Tailed Semi-Supervised Learning}

\begin{document}

\twocolumn[
    \icmltitle{SimPro: A Simple Probabilistic Framework \\ Towards Realistic Long-Tailed Semi-Supervised Learning}

%\icmltitle{Submission and Formatting Instructions for \\
%           International Conference on Machine Learning (ICML 2024)}

% It is OKAY to include author information, even for blind
% submissions: the style file will automatically remove it for you
% unless you've provided the [accepted] option to the icml2024
% package.

% List of affiliations: The first argument should be a (short)
% identifier you will use later to specify author affiliations
% Academic affiliations should list Department, University, City, Region, Country
% Industry affiliations should list Company, City, Region, Country

% You can specify symbols, otherwise they are numbered in order.
% Ideally, you should not use this facility. Affiliations will be numbered
% in order of appearance and this is the preferred way.
\icmlsetsymbol{equal}{*}

\begin{icmlauthorlist}
\icmlauthor{Chaoqun Du}{equal,yyy}
\icmlauthor{Yizeng Han}{equal,yyy}
\icmlauthor{Gao Huang}{yyy}
%\icmlauthor{Firstname4 Lastname4}{sch}
%\icmlauthor{Firstname5 Lastname5}{yyy}
%\icmlauthor{Firstname6 Lastname6}{sch,yyy,comp}
%\icmlauthor{Firstname7 Lastname7}{comp}
%\icmlauthor{}{sch}
%\icmlauthor{Firstname8 Lastname8}{sch}
%\icmlauthor{Firstname8 Lastname8}{yyy,comp}
%\icmlauthor{}{sch}
%\icmlauthor{}{sch}
\end{icmlauthorlist}

\icmlaffiliation{yyy}{Department of Automation, BNRist, Tsinghua University, Beijing, China}
%\icmlaffiliation{comp}{Company Name, Location, Country}
%\icmlaffiliation{sch}{School of ZZZ, Institute of WWW, Location, Country}

%\icmlcorrespondingauthor{Firstname1 Lastname1}{first1.last1@xxx.edu}
\icmlcorrespondingauthor{Gao Huang}{gaohuang@tsinghua.edu.cn}
%\icmlcorrespondingauthor{Firstname2 Lastname2}{first2.last2@www.uk}

% You may provide any keywords that you
% find helpful for describing your paper; these are used to populate
% the "keywords" metadata in the PDF but will not be shown in the document
\icmlkeywords{Machine Learning, ICML}

\vskip 0.3in
]

% this must go after the closing bracket ] following \twocolumn[ ...

% This command actually creates the footnote in the first column
% listing the affiliations and the copyright notice.
% The command takes one argument, which is text to display at the start of the footnote.
% The \icmlEqualContribution command is standard text for equal contribution.
% Remove it (just {}) if you do not need this facility.

%\printAffiliationsAndNotice{}  % leave blank if no need to mention equal contribution
\printAffiliationsAndNotice{\icmlEqualContribution} % otherwise use the standard text.

\begin{abstract}

Recent advancements in semi-supervised learning have focused on a more realistic yet challenging task: addressing imbalances in labeled data while the class distribution of unlabeled data remains both \emph{unknown} and potentially \emph{mismatched}.
Current  approaches in this sphere often presuppose rigid assumptions regarding the class distribution of unlabeled data, thereby limiting the adaptability of models to only certain distribution ranges.
In this study, we propose a novel approach, introducing a highly adaptable framework, designated as \textbf{\name}, which does not rely on any predefined assumptions about the distribution of unlabeled data.
Our framework, grounded in a probabilistic model, innovatively refines the expectation-maximization (EM) algorithm by \emph{explicitly decoupling} the modeling of conditional and marginal class distributions.
This separation facilitates a closed-form solution for class distribution estimation during the maximization phase, leading to the formulation of a Bayes classifier.
The Bayes classifier, in turn, enhances the quality of pseudo-labels in the expectation phase. 
Remarkably, the \name framework not only comes with theoretical guarantees but also is straightforward to implement.
Moreover, we introduce two novel class distributions broadening the scope of the evaluation.
Our method showcases consistent state-of-the-art performance across diverse benchmarks and data distribution scenarios.
Our code is available at \url{https://github.com/LeapLabTHU/SimPro}.

\end{abstract}

\begin{figure}
    \centering
    \includegraphics[width=\linewidth]{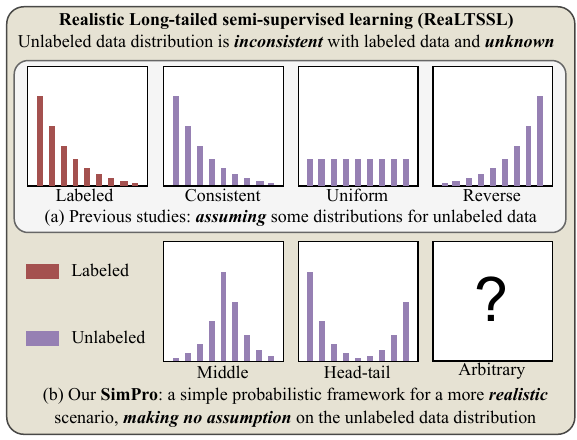}
    \vskip -0.1in
    \caption{The general idea of \name addressing the ReaLTSSL problem. (a) Current methods typically rely on predefined or assumed class distribution patterns for unlabeled data, limiting their applicability. (b) In contrast, our \name embraces a more realistic scenario by introducing a simple and elegant framework that operates effectively without making any assumptions about the distribution of unlabeled data. This paradigm shift allows for greater flexibility and applicability in diverse ReaLTSSL scenarios.}
    \label{fig1:distribution}
    \vskip -0.2in
\end{figure}

\section{Introduction}
\label{sec:intro}

Semi-supervised learning (SSL) offers a viable solution to the scarcity of labeled data by leveraging unlabeled data~\cite{tarvainen2017mean, berthelot2019mixmatch, miyato2018virtual, Sohn2020}.
Common SSL algorithms typically generate pseudo-labels for unlabeled data to facilitate model training~\cite{lee2013pseudo}.
However, real-world data often adheres to a long-tailed distribution,
leading to a predominant focus on majority classes and resulting in imbalanced pseudo-labels~\cite{Liu2019, Kang2020, Du2024}.
This phenomenon, known as long-tailed semi-supervised learning (LTSSL), presents significant challenges in the field.
Traditional LTSSL methods~\cite{lai2022smoothed, lee2021abc, wei2022transfer, Wei2021, Kim2020} assume consistency
in class distributions between labeled and unlabeled data, an often unrealistic premise.
In practice, class distributions can be \emph{inconsistent} and \emph{unknown}, especially as new data are continuously collected or from different tasks.
This ongoing integration process can lead to significant shifts in class distributions.

In response to these challenges, the concept of \emph{realistic long-tailed semi-supervised learning} (ReaLTSSL),
which aims at addressing the \emph{unknown and mismatched} class distributions, has garnered significant attention~\cite{Kim2020,Wei2021,Oh2022, wei2023towards}.
Notably, recent works ACR~\cite{wei2023towards} and  CPE~\cite{Ma2023} pre-define anchor distributions for unlabeled data (\cref{fig1:distribution} (a)).
The ACR estimates the distributional distance to adapt consistency regularization, 
while CPE involves training multiple classifiers, each tailored to a specific class distribution.
However, this approach presupposes certain knowledge about the unlabeled data distribution,
preventing its applications in real-world scenarios where anchor distributions may not represent all possible distributions.
Furthermore, the prevailing techniques often employ multi-branch frameworks and introduce additional loss functions, adding complexity and limiting their generality.

To address these limitations, we propose a \emph{\underline{Sim}ple} \emph{\underline{Pro}babilistic} (\textbf{\name}) framework for ReaLTSSL.
We revisit pseudo-label-based SSL techniques through the lens of the Expectation-Maximization (EM) algorithm.
The EM algorithm, a well-known iterative method in statistical modeling,
is particularly relevant in SSL for handling unobserved latent variables, such as pseudo-labels of unlabeled data.
The E-step entails generating pseudo-labels with the model, while the M-step involves model training using both labeled and unlabeled data.
In the context of \emph{unknown} and \emph{mismatched} class distributions,
the E-step may produce biased pseudo-labels, diminishing the algorithm's effectiveness.
Our \name \emph{avoids fixed assumptions about the unlabeled data distribution}, instead of innovatively extending the EM algorithm for ReaLTSSL.
Specifically, we \emph{explicitly decouple} the modeling of conditional and marginal distributions.
Such separation enables a closed-form solution for the marginal distribution in the M step.
Subsequently, this solution is employed to train a Bayes classifier.
This Bayes classifier, in turn, improves the quality of pseudo-labels generated in the E-step.
Not only does \name offer high effectiveness, but it is also easy to implement, requiring \emph{minimal} code modifications.

Moreover, we expand upon existing evaluation methods~\cite{Oh2022}, which primarily focus on three known class distributions (consistent, uniform, and reversed), by introducing two novel realistic scenarios: middle and head-tail distributions (\cref{fig1:distribution} (b)). The middle distribution represents a concentration of classes in the middle range of labeled data's classes, whereas the head-tail distribution indicates a concentration at both extremes. Notably, our method is theoretically general enough to handle any other distribution patterns, since no prior assumptions are required.

We summarize our contributions as follows:

    1. We present \name, a simple probabilistic framework tailored for realistic long-tailed semi-supervised learning.
    This framework does not presuppose any knowledge about the class distribution of unlabeled data.
    It hinges on the explicit estimation and utilization of class distributions within the EM algorithm.
    \name effectively mitigates the challenges posed by unknown and mismatched class distributions, stepping towards a more realistic LTSSL scenario.
    
    2. We introduce two novel class distribution patterns for unlabeled data, complementing the existing three standard ones.
    This expansion facilitates a more comprehensive and realistic evaluation of ReaLTSSL algorithms,
    bridging the gap between theoretical models and practical applications.
    
    3. Comprehensive experiments on five commonly used benchmarks (CIFAR10/100-LT,  STL10-LT, and ImageNet-127/1k)
and five distinct class distributions validate that our \name consistently achieves SOTA performance.

\section{Related Work}
\label{sec:related}

\textbf{Semi-supervised learning} (SSL) has gained prominence through a subset of algorithms that use unlabeled data to enhance model performance. This enhancement primarily occurs through the generation of pseudo-labels, effectively forming a self-training loop \cite{miyato2018virtual,berthelot2019remixmatch,berthelot2019mixmatch,huang2022high,Wang2023}. Modern SSL methodologies, such as those presented in \cite{berthelot2019remixmatch,Sohn2020}, integrate pseudo-labeling with consistency regularization. This integration fosters uniform predictions across varying representations of a single image, thereby bolstering the robustness of deep networks. A notable example, FixMatch \cite{Sohn2020}, has demonstrated exceptional results in image recognition tasks, outperforming competing SSL approaches. 

The efficacy of SSL algorithms heavily relies on the quality of the pseudo-labels they generate. However, both labeled and unlabeled data follow a long-tailed class distribution in the LTSSL scenario. Conventional SSL methods are prone to produce biased pseudo-labels, which significantly downgrade their effectiveness.

\noindent\textbf{Long-tailed semi-supervised learning} has garnered considerable interest due to its relevance in numerous real-world applications.
In this domain, DARP \cite{Kim2020} and CReST \cite{Wei2021} aim to mitigate the issue of biased pseudo-labels by aligning them with the class distribution of labeled data. Another notable approach \cite{lee2021abc} employs an auxiliary balanced classifier, which is trained through the down-sampling of majority classes, to enhance generalization capabilities. 
These algorithms have markedly improved performance but operate under the assumption of identical class distributions for labeled and unlabeled data. 

In addressing Realistic LTSSL challenges, DASO~\cite{Oh2022} innovates by adapting the proportion of linear and semantic pseudo-labels to the unknown class distribution of unlabeled data.
Its success largely depends on the discriminative quality of the representations, a factor that becomes less reliable in long-tailed distributions.
ACR~\cite{wei2023towards}, on the other hand, attempts to refine consistency regularization by pre-defining distribution anchors and achieves promising results.
CPE~\cite{Ma2023} trains multiple anchor experts where each is tasked to model one distribution.
However, such anchor distribution-based approaches might not encompass all potential class distribution scenarios, and their complexity could hinder the broader application.

\section{Method}\label{sec:method}
In this section, we first introduce the problem formulation of ReaLTSSL (\cref{sec:preliminary}), setting the stage for our method. Subsequently, we delve into the proposed simple and probabilistic framework, \name (\cref{sec:framework}). We provide implementation details in \cref{sec:implementation} to elucidate \name further.

\subsection{Preliminaries}\label{sec:preliminary}
\paragraph{Problem formulation.}
We begin by outlining the formulation for the realistic long-tailed semi-supervised learning (ReaLTSSL) problem, laying the groundwork for our approach.
The setup involves a labeled dataset $\mathcal{D}_l\!=\!\{(x_i, y_i)\}_{i=1}^{N}$ and an unlabeled dataset $\mathcal{D}_u\!=\!\{x_i\}_{i=1}^{M}$, where $x_i\!\in\!\mathbb{R}^d$ represents the $i$-th data sample and $y_i\!\in\!\{0,1\}^K$ is the corresponding one-hot label, with $K$ denoting the number of classes.
The objective of ReaLTSSL is to train a classifier $F_{\bm{\theta}}: \mathbb{R}^d\!\mapsto\!\{0,1\}^K$, parameterized by $\bm{\theta}$.

\begin{assumption}
    \label{assumption}
We assume a realistic scenario where labeled, unlabeled, and test data share the same conditional distribution $P(x|y)$,
yet may exhibit distinct marginal distributions $P(y)$.
Crucially, the marginal distribution $P(y)$ of the unlabeled data remains \emph{unknown}. 
\end{assumption}
Further, we consider five diverse distributions for the unlabeled data (\cref{fig1:distribution}), reflecting various real-world situations.

\paragraph{The EM algorithm in semi-supervised learning.}
In SSL, pseudo-labeling is a key technique for leveraging unlabeled data.
This involves creating pseudo-labels for the unlabeled data using the model and then training the model with both the pseudo-labeled and ground-truth data.
This aligns with the Expectation-Maximization (EM) algorithm~\cite{dempster1977maximum}, where the E-step generates pseudo-labels, and the M-step updates the parameters using the pseudo-labels, maximizing the likelihood function.

Our method builds on a popular algorithm FixMatch~\cite{Sohn2020}, which integrates consistency regularization in the standard SSL setting. Pseudo-labels are created via \emph{weakly-augmented} unlabeled data and applied to train \emph{strongly-augmented} samples based on a confidence threshold. The loss for unlabeled data is
\begin{equation}\label{eq:FixMatch}
    \mathcal{L}_{u}(x_i) = \mathbb{I}(\text{max}(q_{\omega})\geq t)\cdot\mathcal{H}(\arg\max(q_{\omega}),q_{\Omega}),
\end{equation}
where $q_{\omega}$ and $q_{\Omega}$ represent the prediction logits for weakly and strongly augmented samples, respectively, $\mathcal{H}$ denotes the cross-entropy loss, and $t$ is the confidence threshold.

\paragraph{Long-tailed learning.}
In typical SSL scenarios, the assumption of identical distributions for labeled, unlabeled, and test data often prevails. However, long-tailed learning tasks usually involve imbalanced training sets and balanced test sets, leading to discrepancies in the prior distribution of $P(y)$ between training and testing data. Some studies \cite{Ren2020,Menon2021,Hong2021} tackle this via Bayesian inference, introducing a prior distribution over class labels:
\begin{align}\label{eq:BC}
\mathcal{L}_l(x) &= -\log P(y|x;\bm{\theta},\bm{\pi}) \nonumber\\
&= -\log \frac{P(y;\bm{\pi})P(x|y;\bm{\theta})}{P(x)}  \nonumber\\
&= -\log \frac{\phi_y\exp(f_{\bm{\theta}}(x,y))}{\sum_{y'}\phi_{y'}\exp(f_{\bm{\theta}}(x,y'))},
\end{align}
where $\phi_y$ denotes the class frequency in the training or test set,  $\bm{\pi}$ is the class distribution parameter and $\bm{\theta}$ is the parameter of $P(x|y)/P(x)$.
Here we omit the parameter of $P(x)$ for simplicity.
The detailed mathematical derivation is provided in \cref{sec:detailed_model}.

While supervised learning allows for a known distribution parameter $\bm{\pi}$, enabling a direct application to model $P(y)$ and explicit decoupling from $\bm{\theta}$, ReaLTSSL poses a greater challenge as the prior $\bm{\pi}$ for unlabeled data is \emph{unknown}. This necessitates innovative approaches to adapt to the imbalanced data while maintaining model efficacy.

\begin{figure*}
    \centering
    \includegraphics[width=0.8\linewidth]{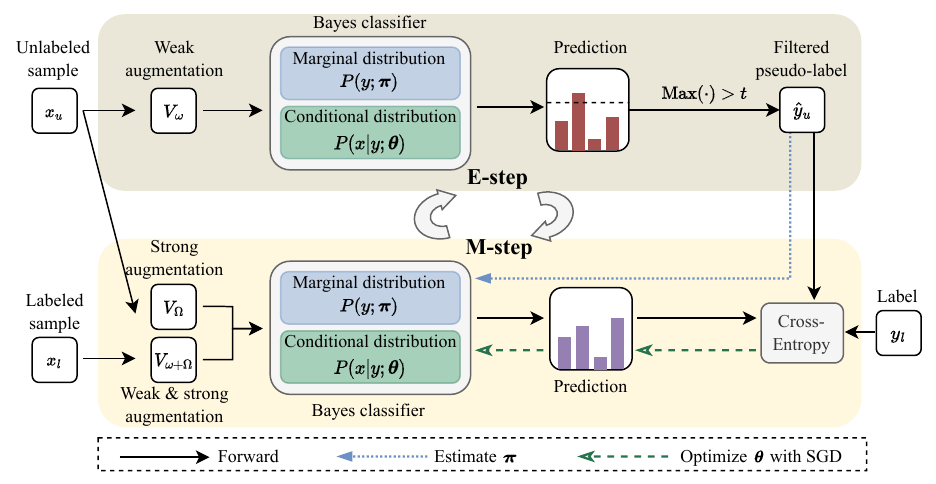}
    \caption{The \name Framework Overview. This framework distinctively separates the conditional and marginal (class) distributions. In the E-step (top), pseudo-labels are generated using the current parameters $\bm{\theta}$ and $\bm{\pi}$.  In the subsequent M-step (bottom), these pseudo-labels, along with the ground-truth labels, are utilized to compute the Cross-Entropy loss (refer to \cref{eq:overall_loss}), facilitating the optimization of network parameters $\bm{\theta}$ via gradient descent. Concurrently, the marginal distribution parameter $\bm{\pi}$ is recalculated using a closed-form solution based on the generated pseudo-labels (as detailed in \cref{eq:pi}).}
    \label{fig:framework}
\end{figure*}
\subsection{\name Framework}\label{sec:framework}

\paragraph{Framework overview.}
In the realistic semi-supervised learning (ReaLTSSL) context, the conventional assumption of independent and identically distributed (i.i.d.) labeled and unlabeled data is no longer valid. Moreover, the marginal (class) distribution $P(y)$ of the unlabeled data may be \emph{inconsistent} with that of the labeled data and remains \emph{unknown}, which challenges the traditional SSL frameworks.

To overcome this, we introduce \name, an elegant and effective probabilistic framework adapted for the unique ReaLTSSL setting.
Illustrated in \cref{fig:framework}, \name distinctively decouples $\bm{\pi}$ and $\bm{\theta}$, unlike traditional SSL methods \cite{Sohn2020}.
In the E-step, we generate pseudo-labels using the parameters $\bm{\pi}$ and $\bm{\theta}$ obtained from the previous M-step. The M-step then models the conditional distribution $P(x|y)$ using network parameters $\bm{\theta}$, which are optimized through gradient descent. Simultaneously, we derive a closed-form solution for the class distribution $P(y)$, represented by $\bm{\pi}$.

It is worth noting that the treatment of the $\bm{\pi}$ in our framework is not heuristic. It is firmly rooted in probabilistic modeling and the principles of the EM algorithm, providing theoretical soundness as substantiated in \cref{prop:pi,prop:phi}.

\paragraph{Probabilistic model.}
In addressing the ReaLTSSL challenge, we adopt an Expectation-Maximization (EM) approach, underpinned by a robust probabilistic model.
The model is governed by the fundamental principles of conditional probability, as shown in:
\begin{equation}
    P(\bm{y},\bm{x};\bm{\theta},\bm{\pi}) = P(\bm{y}|\bm{x};\bm{\theta},\bm{\pi})P(\bm{x}).
    \label{eq:prob}
\end{equation}
Here, we do not explicitly parameterize $P({x})$, as per the independence of parameters through conditional parameterization~\cite{Koller2009}.
Thus, when ${x}$ is not a condition, the parameters of the relevant notions omit the parameters of $P({x})$, such as $P({x})$, $P({x}|y)$, $P({x},y)$, etc.
According to \cref{eq:BC}, this may lead to a potential misunderstanding, as the equation $P({x}) = \sum_{y}P({x}|y;\bm{\theta})P(y;\bm{\pi})$ seems to suggest that $P({x})$ is parameterized by $\bm{\theta}$ and $\bm{\pi}$, which is not the case.
The detailed mathematical derivation is provided in \cref{sec:detailed_model}.

We focus on estimating the parameters $\bm{\theta}$ and $\bm{\pi}$, pivotal for learning a discriminative model.
Consequently, we concentrate on those terms dependent on $\bm{\theta}$ and $\bm{\pi}$, sidelining those independent of these parameters.

The complete data log-likelihood is thus articulated as:
\begin{equation}
    \small
    P(\bm{y}|\bm{x};\bm{\theta},\bm{\pi}) = \prod_{i=1}^{N}P(y_i|x_i;\bm{\theta},\bm{\pi}_l) \prod_{j=1}^{M}P(y_j|x_j;\bm{\theta},\bm{\pi}_u),
    \label{eq:complete}
\end{equation}
where $\bm{\pi}\!=\!\{\bm{\pi}_l,\bm{\pi}_u\}$ signifies the class distributions for labeled and unlabeled data, respectively, with $N$ and $M$ representing the number of labeled/unlabeled samples.

\paragraph{E-step (generating pseudo-labels).} 
By \cref{eq:complete}, the expected complete data log-likelihood $\mathcal{Q}$ function is derived from the preceding iteration's parameters, $\bm{\theta}'$ and $\bm{\pi}'$:
\begin{align}
    \mathcal{Q}(\bm{\theta},\bm{\pi};\bm{\theta}',\bm{\pi}') &= \mathbb{E}_{\bm{y}|\bm{x};\bm{\theta}',\bm{\pi}'}\left[\log P(\bm{y},\bm{x};\bm{\theta},\bm{\pi})\right] \label{eq:log_likelihood} \\
    &= \sum_{i}\log P(y_i|x_i;\bm{\theta},\bm{\pi}_l) \nonumber\\
    &+ \sum_{j,y}P(y|x_j;\bm{\theta}',\bm{\pi}')\log P(y|x_j;\bm{\theta},\bm{\pi}_u). \nonumber
\end{align}
The E-step involves generating soft pseudo-labels $P(y|x_j;\bm{\theta}',\bm{\pi}')$ under the current $\bm{\theta}'$ and $\bm{\pi}'$.
These soft pseudo-labels are specifically defined by \cref{eq:BOC}, which is detailed in \cref{prop:phi}.
In the subsequent M-step, these pseudo-labels are used alongside the one-hot labels of the labeled data to compute the cross-entropy loss.

\paragraph{M-step (optimizing $\bm{\theta}$ and $\bm{\pi}$).}
The M-step focuses on optimizing the expected complete data log-likelihood $\mathcal{Q}$-function concerning the parameters $\bm{\theta}$ and $\bm{\pi}$.

\noindent (a) \emph{Optimization of $\bm{\pi}$}: The closed-form solution for $\bm{\pi}$ can be derived directly from the $\mathcal{Q}$-function (\cref{eq:log_likelihood}).
Specifically, the terms involving $\bm{\pi}$ in $\mathcal{Q}(\bm{\theta},\bm{\pi};\bm{\theta}',\bm{\pi}')$ are given by
\begin{equation}
 \sum_{i}\log P(y_i;\bm{\pi}_l) + \sum_{j,y}P(y|x_j;\bm{\theta}',\bm{\pi}')\log P(y;\bm{\pi}_u).
\end{equation}

\begin{proposition}[Closed-form Solution for $\bm{\pi}$]
\label{prop:pi}
The optimal $\bm{\hat\pi}$ that maximizes $\mathcal{Q}(\bm{\theta},\bm{\pi};\bm{\theta}',\bm{\pi}')$ is
\begin{equation}
    \bm{\hat\pi}_l = \frac{1}{N}\sum_{i=1}^{N}{y_i}, \quad \bm{\hat\pi}_u = \frac{1}{M}\sum_{j=1}^{M}P(y|x_j;\bm{\theta}',\bm{\pi}').
    \label{eq:pi}
\end{equation}
\end{proposition}
\vspace{-5pt}

\noindent (b) \emph{Optimization of $\bm{\theta}$}: 
The network parameters $\bm{\theta}$, unlike $\bm{\pi}$ which have a closed-form solution, are optimized using standard stochastic gradient descent (SGD). 
Combining with \cref{eq:BC}, the terms involving $\bm{\theta}$ in $\mathcal{Q}(\bm{\theta},\bm{\pi};\bm{\theta}',\bm{\pi}')$ are
\begin{align}
    &(\sum_{i} + \sum_{j,y}P(y|x_j;\bm{\theta}',\bm{\pi}')) \log\frac{P(x|y;\bm{\theta})}{P(x)}  \label{eq:theta} \\
   =&(\sum_{i} + \sum_{j,y}P(y|x_j;\bm{\theta}',\bm{\pi}')) \log \frac{\exp(f_{\bm{\theta}}(x,y))}{\sum_{y'}\phi_{y'}\exp(f_{\bm{\theta}}(x,y'))}, \nonumber 
\end{align}
which simplifies to the supervised scenario in \cref{eq:BC} by treating $P(y|x_j;\bm{\theta}',\bm{\pi}')$ as soft labels.
Maximizing \cref{eq:theta} corresponds to minimizing the cross-entropy loss.
Here, $\phi_{y'}$ is interpreted as the estimated overall frequency of class $y'$. 
The optimization of model parameters $\bm{\theta}$ using the overall frequency vector $\bm{\phi}$ is crucial for learning a Bayes classifier.

\begin{proposition}[Bayes Classifier]
\label{prop:phi}
In conjunction with the high-confidence filtering (\cref{eq:FixMatch}),
the optimal $\bm{\hat\phi}$ for learning a Bayes classifier is mathematically derived as:
\begin{align}
    \bm{\hat\phi} &= [\hat\phi_1,\hat\phi_2,\cdots,\hat\phi_K] \nonumber\\
                 &= \frac{1}{N+M}(\sum_i y_i + \sum_{j}P(y|x_j;\bm{\theta}',\bm{\pi}')).
    \label{eq:phi}
\end{align}
Subsequently, with the model parameter $\bm{\theta}$  which is optimized using the $\bm{\hat\phi}$,
the corresponding Bayes classifier for unlabeled or test dataset with estimated or uniform class distribution is defined by the equation:
\begin{align}
    P(y|x;\bm{\theta};\bm{\hat\pi}) & = \frac{P(y;\bm{\hat\pi})\exp(f_{\bm{\theta}}(x,y))}{\sum_{y'}P(y';\bm{\hat\pi})\exp(f_{\bm{\theta}}(x,y'))}, \label{eq:BOC} \\
    \text{or} \quad P(y|x;\bm{\theta}) & = \frac{\exp(f_{\bm{\theta}}(x,y))}{\sum_{y'}\exp(f_{\bm{\theta}}(x,y'))}. \label{eq:BOC_uniform}
\end{align}
\end{proposition}
\vspace{-5pt}

Building upon \cref{prop:phi}, it is crucial to acknowledge that the parameter vector $\bm{\phi}$ is vital for learning Bayes classifiers.
Consequently, to delve deeper into the theoretical foundations, we evaluate the impact of $\bm{\phi}$ on the model’s performance.
In line with the principles of online decision theory, we establish a regret bound for the decision error rate on the test set, denoted as $P(e;\bm\phi)$.
Our analysis is simplified by concentrating on a binary classification scenario, where the labels $y$ belong to $\{-1, +1\}$.

\begin{proposition}[Regret Bound]
\label{prop:regret}
Let $\bm{\phi}^*$ denote the vector $\bm{\phi}$ obtained in \cref{eq:phi} when pseudo-labels are replaced by ground-truth labels.
For the decision error rate $P(e;\bm\phi)$ on the test set, the regret bound is expressed as:
    \begin{equation}
        P(e;\bm{\hat\phi})-\inf_{\bm\phi} P(e;\bm{\phi}) \le \frac{1}{2\phi^*_{+1}\phi^*_{-1}} |\hat\phi - \phi^*|,
    \end{equation}
where $|\hat\phi - \phi^*| = |\hat\phi_{+1} - \phi^*_{+1}| = |\hat\phi_{-1} - \phi^*_{-1}|$.
\end{proposition}
\vspace{-5pt}
\cref{prop:regret}~illustrates that the regret bound is primarily governed by the first-order term of the estimation deviation.
Additionally, it is inversely proportional to the ground truth $\bm{\phi}^*$,
highlighting the learning challenges associated with imbalanced training data from a regret-bound perspective.

\begin{algorithm}[t]
\caption{Pseudocode of \name in a PyTorch-like style.}
\label{alg:code}
\definecolor{codeblue}{rgb}{0.25,0.5,0.5}
\lstset{
  backgroundcolor=\color{white},
  basicstyle=\fontsize{7.2pt}{7.2pt}\ttfamily\selectfont,
  columns=fullflexible,
  breaklines=true,
  captionpos=b,
  commentstyle=\fontsize{7.2pt}{7.2pt}\color{codeblue},
  keywordstyle=\fontsize{7.2pt}{7.2pt},
  escapeinside={(*@}{@*)},
}
\begin{lstlisting}[language=python]
# N_e: (K,), where N_e[k] denotes the number of labeled data in class k in one epoch
# pi_u: (K,), the class distribution parameters of unlabeled samples
# phi: (K,), the overall class frequency
# f: deep network parameterized by theta
# alpha, tau, m: hyper-parameters
# CE: CrossEntropyLoss
# aug_w, aug_s: weak and strong augmentation
pi_u.init(uniform)
phi.init(consistent)
for epoch in range(epochs):  
    pi_e = zeros(K) # temporary estimation of pi_u   
    # load labeled and unlabeled samples
    for (x_l, y_l), x_u in zip(loader_l, loader_u):
        
        # E step: generating pseudo labels
        lgt_l = f.forward(aug(x_l)) 
        lgt_w = f.forward(aug_w(x_u)).detach()
        lgt_s = f.forward(aug_s(x_u)) 
        # Bayes classifer
        psd_lbs = softmax(lgt_w + (*@\textbf{tau*log(pi\_u)}@*), dim=-1)
        # filter out pseudo labels with high confidence
        mask = max(psd_lbs, dim=-1)[0].ge(t) 

        # M step: solving pi and phi, optimizing theta
        # solve pi_u with Eq. (7)
        (*@\textbf{pi\_e += sum(psd\_lbs[mask], dim=0)}@*) 
        # optimize f (theta) with Eq. (11)
        loss_l = CE(lgt_l + (*@\textbf{tau * log(phi)}@*), y_l)
        loss_u = mean(CE(lgt_s + (*@\textbf{tau * log(phi)}@*), 
                  psd_lbs, reduction='none') * mask)
        loss = (*@\textbf{alpha}@*) * loss_l + loss_u
        loss.backward()
        update(theta)
    
    # update pi_u and phi
    phi_e = (pi_e + N_e) / sum(pi_e + N_e) # Eq. (9)
    # moving average
    phi = m * phi + (1 - m) * phi_e
    pi_u = m * pi_u + (1 - m) * pi_e / sum(pi_e)
\end{lstlisting}
\end{algorithm}

\begin{table*}[!t]
\caption{Top-1 accuracy ($\%$) on CIFAR10-LT ($N_1=500,M_1=4000$) with different class imbalance ratios $\gamma_l$ and $\gamma_u$ under five different unlabeled class distributions.
$\dagger$ indicates we reproduce ACR without anchor distributions for a fair comparison.}
\label{tab:main_cifar10}
\centering
 \resizebox{1.0\linewidth}{!}{%
     \begin{tabular}{lcccccccccc}
         \toprule
            & \multicolumn{2}{c}{consistent} & \multicolumn{2}{c}{uniform}  & \multicolumn{2}{c}{reversed} & \multicolumn{2}{c}{middle} & \multicolumn{2}{c}{head-tail} \\
        \cmidrule(lr){2-3} \cmidrule(lr){4-5} \cmidrule(lr){6-7} \cmidrule(lr){8-9} \cmidrule(l){10-11}
          & \multicolumn{1}{c}{$\gamma_l=150$} & \multicolumn{1}{c}{$\gamma_l=100$} & \multicolumn{1}{c}{$\gamma_l=150$} & \multicolumn{1}{c}{$\gamma_l=100$} & \multicolumn{1}{c}{$\gamma_l=150$} & \multicolumn{1}{c}{$\gamma_l=100$} & \multicolumn{1}{c}{$\gamma_l=150$} & \multicolumn{1}{c}{$\gamma_l=100$}  & \multicolumn{1}{c}{$\gamma_l=150$} & \multicolumn{1}{c}{$\gamma_l=100$} \\
        \cmidrule(lr){2-3} \cmidrule(lr){4-5} \cmidrule(lr){6-7} \cmidrule(lr){8-9} \cmidrule(l){10-11}
          & \multicolumn{1}{c}{$\gamma_u=150$} & \multicolumn{1}{c}{$\gamma_u=100$} & \multicolumn{1}{c}{$\gamma_u=1$} & \multicolumn{1}{c}{$\gamma_u=1$} & \multicolumn{1}{c}{$\gamma_u=1/150$} & \multicolumn{1}{c}{$\gamma_u=1/100$} & \multicolumn{1}{c}{$\gamma_u=150$} & \multicolumn{1}{c}{$\gamma_u=100$}  & \multicolumn{1}{c}{$\gamma_u=150$} & \multicolumn{1}{c}{$\gamma_u=100$} \\
       \cmidrule(r){1-1}  \cmidrule(lr){2-3} \cmidrule(lr){4-5} \cmidrule(lr){6-7} \cmidrule(lr){8-9} \cmidrule(l){10-11}
         FixMatch~\cite{Sohn2020}   & \ms{62.9}{0.36} & \ms{67.8}{1.13}& \ms{67.6}{2.56}     & \ms{73.0}{3.81}  & \ms{59.9}{0.82}     & \ms{62.5}{0.94} & \ms{64.3}{0.63} & \ms{71.7}{0.46}     & \ms{58.3}{1.46} & \ms{66.6}{0.87}    \\
        ~~w/ CReST+~\cite{Wei2021} & \ms{67.5}{0.45} & \ms{76.3}{0.86}& \ms{74.9}{0.80}     & \ms{82.2}{1.53}  & \ms{62.0}{1.18}     & \ms{62.9}{1.39} & \ms{58.5}{0.68} & \ms{71.4}{0.60}     & \ms{59.3}{0.72} & \ms{67.2}{0.48}    \\
        ~~w/ DASO~\cite{Oh2022}   & \ms{70.1}{1.81} & \ms{76.0}{0.37}& \ms{83.1}{0.47}     & \ms{86.6}{0.84}  & \ms{64.0}{0.11}     & \ms{71.0}{0.95} & \ms{69.0}{0.31} & \ms{73.1}{0.68}     & \ms{70.5}{0.59} & \ms{71.1}{0.32}    \\
        ~~w/ ACR$^\dagger$~\cite{wei2023towards}    & \ms{70.9}{0.37} & \ms{76.1}{0.42}& \ms{91.9}{0.02} & \ms{92.5}{0.19}  & \ms{83.2}{0.39} & \ms{85.2}{0.12} & \ms{73.8}{0.83} & \ms{79.3}{0.30} & \ms{77.6}{0.20} & \ms{79.3}{0.48}\\
       \cmidrule(r){1-1} \cmidrule{2-3} \cmidrule(lr){4-5} \cmidrule(lr){6-7} \cmidrule(lr){8-9} \cmidrule(l){10-11}
\rowcolor{mygray}
~~w/ \name & \msb{74.2}{0.90} & \msb{80.7}{0.30} & \msb{93.6}{0.08} & \msb{93.8}{0.10} & \msb{83.5}{0.95} & \msb{85.8}{0.48} & \msb{82.6}{0.38} & \msb{84.8}{0.54} & \msb{81.0}{0.27} & \msb{83.0}{0.36}  \\
        \bottomrule
    \end{tabular}
}
\vskip -0.1in
\end{table*}

\begin{table*}[t]
\caption{Top-1 accuracy (\%) on CIFAR100-LT and STL10-LT with different class imbalance ratios $\gamma_l$ and $\gamma_u$.
Due to the unknown ground-truth labels of the unlabeled data for STL10, we conduct the experiments by controlling the imbalance ratio of the labeled data.
$\dagger$ indicates we reproduce the results of ACR without anchor distributions for fair comparison.}
\label{tab:main_cifar100}
\centering
\resizebox{1.0\linewidth}{!}{%
    \begin{tabular}{lccccclcc}
        \toprule
         & \multicolumn{5}{c}{CIFAR100-LT ($\gamma_l=20, N_1=50, M_1=400$) } & & \multicolumn{2}{c}{STL10-LT ($\gamma_u=\emph{N/A}$)} \\
         \cmidrule(lr){2-6}  \cmidrule(l){8-9}
         & \multicolumn{1}{c}{$\gamma_u=20$} & \multicolumn{1}{c}{$\gamma_u=1$}  & \multicolumn{1}{c}{$\gamma_u=1/20$} & \multicolumn{1}{c}{$\gamma_u=20$} & \multicolumn{1}{c}{$\gamma_u=20$} & & \multicolumn{2}{c}{$N_1=450, \quad M=1\!\times\!10^5$} \\
         \cmidrule(lr){2-6} \cmidrule(l){8-9}
         & \multicolumn{1}{c}{consistent} & \multicolumn{1}{c}{uniform}  & \multicolumn{1}{c}{reversed} & \multicolumn{1}{c}{middle} & \multicolumn{1}{c}{head-tail} &  & \multicolumn{1}{c}{$\gamma_l=10$} & \multicolumn{1}{c}{$\gamma_l=20$} \\
         \cmidrule(r){1-1}  \cmidrule(lr){2-6}   \cmidrule(lr){7-7} \cmidrule(l){8-9}

        FixMatch~\cite{Sohn2020} & \ms{40.0}{0.96}  & \ms{39.6}{1.16}  & \ms{36.2}{0.63} & \ms{39.7}{0.61} & \ms{38.2}{0.82} & FixMatch~\cite{Sohn2020}  &\ms{72.4}{0.71} & \ms{64.0}{2.27} \\
        ~~w/ CReST+~\cite{Wei2021}  & \ms{40.1}{1.28} & \ms{37.6}{0.88} & \ms{32.4}{0.08} & \ms{36.9}{0.57} & \ms{35.1}{1.10} & ~~w/ CReST+~\cite{Wei2021}  & \ms{71.5}{0.96} & \ms{68.5}{1.88} \\
        ~~w/ DASO~\cite{Oh2022}                   & \ms{43.0}{0.15} & \ms{49.4}{0.93} & \ms{44.1}{0.25} & \ms{43.1}{1.20} & \ms{43.8}{0.43} & ~~w/ DASO~\cite{Oh2022}  & \ms{78.4}{0.80} & \ms{75.3}{0.44} \\
        ~~w/ ACR$^\dagger$~\cite{wei2023towards}         & \ms{40.7}{0.57} & \ms{50.2}{0.82} &   \ms{44.1}{0.14} &  \ms{42.4}{0.47} & \ms{41.1}{0.09} &  ~~w/ ACR~\cite{wei2023towards}     & \ms{83.0}{0.32} & \ms{81.5}{0.25} \\
        \cmidrule(r){1-1}  \cmidrule(lr){2-6}   \cmidrule(lr){7-7} \cmidrule(l){8-9}
\rowcolor{mygray}
        ~~w/ \name                       & \msb{43.1}{0.40} & \msb{52.2}{0.16} & \msb{45.5}{0.34} & \msb{43.6}{0.35} & \msb{44.8}{0.56} & ~~w/ \name & \msb{84.5}{0.39} & \msb{82.5}{0.25} \\

        \bottomrule

    \end{tabular}
}
\vskip -0.10in
\end{table*}

\subsection{Implementation Details}\label{sec:implementation}

\paragraph{Training objective for optimizing $\bm{\theta}$.} 
In \name, the E-step primarily involves generating pseudo-labels using parameters $\bm{\theta}$ and $\bm{\pi}$. Consequently, in the M-step, we first focus on optimizing the network parameter $\bm{\theta}$ guided by \cref{eq:theta} via Stochastic Gradient Descent (SGD). Building on the FixMatch algorithm~\cite{Sohn2020}, the overall training objective is formulated as:
\begin{equation}\label{eq:overall_loss}
    \mathcal{L} = \alpha\mathcal{L}_l + \mathcal{L}_u,
\end{equation}
where $\mathcal{L}_l$ and $\mathcal{L}_u$ represent the losses on labeled and unlabeled data, respectively.
The hyper-parameter $\alpha$ acts as a scaling factor, the specifics of which are elucidated later. 

For $\mathcal{L}_l$, we modify it (originally the standard cross-entropy loss) following \cref{eq:theta}:
\begin{equation}
    \mathcal{L}_l = - \frac{1}{B}\sum_{i=1}^{B}\log\frac{\exp(f_{\bm{\theta}}(x_i,y_i))}{\sum_{y'}\phi_{y'}^\tau \exp (f_{\bm{\theta}}(x_i,y'))},
    \label{eq:l_l}
\end{equation}
where $\tau$ is a hyper-parameter for enhancing adaptability to long-tail distributions~\cite{Menon2021}. $B$ is batch size.

For $\mathcal{L}_u$, we implement the standard SSL format (\cref{eq:FixMatch}) and adapt it for ReaLTSSL:
\begin{align}\label{eq:l_u}
    \mathcal{L}_u = - \frac{1}{\mu B}\sum_{j=1}^{\mu B}\mathbb{I}(\text{max}_{y}(q_y))\geq t)\sum_{y}q_{y} \log p_y,
\end{align}
where $\mu$ controls the number of unlabeled samples, and $t$ is a confidence threshold.
The pseudo-label of weak augmentation $\omega$ from the Bayes classifier (\cref{eq:BOC}) is denoted by
\begin{equation}
    q_y = P(y|\omega(x_j);\bm{\theta},\bm{\hat\pi}),
\end{equation}
and the actual prediction  $p_y$ is obtained using strong augmentation $\Omega$ and calibrated with $\bm{\phi}$ as shown in \cref{eq:theta}:
\begin{equation}
    p_y = \frac{\exp(f_{\bm{\theta}}(\Omega(x_j),y))}{\sum_{y'}\phi_{y'}^\tau \exp (f_{\bm{\theta}}(\Omega(x_j),y'))}.
\end{equation}

Moreover, in practical situations, the size of the unlabeled dataset $M$ is generally larger than that of the labeled dataset $N$.
To ensure a balanced sample size in each iteration, we usually set $\mu\!=\!M/N$ in \cref{eq:l_u}.
In specific scenarios, we further adjust the balance factor in \cref{eq:overall_loss} to $\alpha\!=\!\mu\cdot N/M\!<\!1$. 
This methodology effectively mitigates overfitting to the labeled data (see \cref{tab:recipe,tab:alpha}).
\paragraph{Closed-form solution of $\bm{\pi}$.} As discussed in \cref{sec:framework},
the parameter $\bm{\pi}$ of marginal distribution $P(y)$,
has a closed-form solution in the M-step.
Therefore, unlike $\bm{\theta}$, which requires SGD optimization, $\bm{\pi}$ (\cref{eq:pi}) and $\bm{\phi}$ (\cref{eq:phi}) are computed and updated via a moving average during training.

\paragraph{Extended EM algorithm and pseudo-code.} Based on the previous analysis, Our \name framework can be summarized as an extended EM algorithm, which includes:

   $\bm{\cdot}$ \emph{E-step} (\cref{eq:BOC}): Generating pseudo-labels using model parameters $\bm{\theta}$ and estimated distribution parameters $\bm{\pi}$;
   
   $\bm{\cdot}$\ \emph{M-step}: Optimizing network parameters $\bm{\theta}$ via SGD using \cref{eq:theta} (actually using \cref{eq:overall_loss}), and solving distribution parameters $\bm{\pi}$ and hyper-parameters $\bm{\phi}$ by \cref{eq:pi} and \cref{eq:phi}.

For further clarity, the pseudo-code of \name is provided in \cref{alg:code}.
The modifications we made to the core training code, in comparison to FixMatch, are highlighted in bold. 
In the \name, we incorporate just a \emph{single additional line} of code in the M-step to compute the closed-form solution of $\pi$ (\cref{prop:pi}).
Furthermore, only \emph{four lines} of code need to be modified to construct a Bayes classifier (\cref{prop:phi}) and to balance the loss between labeled and unlabeled data (denoted as $\alpha$).
These minor yet crucial adjustments demonstrate that our \name framework is not only grounded in \emph{rigorous theoretical derivation} but is also straightforward to implement in practice, exemplifying both \emph{simplicity and elegance}.

\section{Experiments}
\label{sec:exp}

In this section, we first present the main results on various ReaLTSSL benchmarks in \cref{sec:main_results}. 
More analysis, including the ablation studies and the visualization results, is presented in \cref{sec:analysis} to further evaluate the effectiveness of our \name.
For detailed information regarding the experimental setup, please refer to \cref{sec:setup}.

\begin{table}[t]
\caption{ Top-1 accuracy ($\%$) on ImageNet-127 ($\gamma_l\!=\!\gamma_u\approx286, N_1\approx28\text{k}$, and $M_1\approx250\text{k}$)
    and ImageNet-1k ($\gamma_l\!=\!\gamma_u\!=256, N_1\!=\!256, M_1\!=\!1024$)
    with different test class imbalance ratios $\gamma_t$ and image resolutions.
$\dagger$ indicates we reproduce ACR without anchor distributions for a fair comparison.
    The results of $\gamma_t\approx286$ are sourced from ACR~\cite{wei2023towards}.
}
\label{tab:main_imagenet}

\centering
\resizebox{0.8\linewidth}{!}{%
    {
    \begin{tabular}{lcc}
        \toprule

        & \multicolumn{2}{c}{$\gamma_t\approx 286$} \\
         \cmidrule(l){2-3}
         \textit{ImageNet-127}  & $32\times32$ & $64\times64$ \\
         \cmidrule(r){1-1} \cmidrule(l){2-3}

         FixMatch~\cite{Sohn2020} & 29.7 & 42.3 \\
         ~~w/ DARP~\cite{Kim2020} & 30.5 & 42.5 \\
         ~~w/ CReST+~\cite{Wei2021}  & 32.5 & 44.7 \\
         ~~w/ CoSSL~\cite{Fan2022}  & 43.7 & 53.9 \\
         ~~w/ ACR~\cite{wei2023towards} & 57.2 & 63.6 \\
        \cmidrule(r){1-1} \cmidrule(l){2-3} 
\rowcolor{mygray}
         ~~w/ \name                       & \textbf{59.1} & \textbf{67.0} \\

         \midrule
        \textit{ImageNet-127}  & \multicolumn{2}{c}{$\gamma_t = 1 $} \\
         \cmidrule(r){1-1} \cmidrule(l){2-3}

         FixMatch~\cite{Sohn2020} & 38.7 & 46.7 \\
         ~~w/ ACR$^\dagger$~\cite{wei2023towards}   & 49.5 &  56.1 \\
         ~~w/ ACR~\cite{wei2023towards} & 50.6 & 57.3  \\
        \cmidrule(r){1-1} \cmidrule(l){2-3}
\rowcolor{mygray}
         ~~w/ \name & \textbf{55.7} & \textbf{63.8} \\

         \midrule

        \textit{ImageNet-1k} & \multicolumn{2}{c}{$\gamma_t = 1 $} \\
         \cmidrule(r){1-1} \cmidrule(l){2-3}

         FixMatch~\cite{Sohn2020} & -- & -- \\
         ~~w/ ACR$^\dagger$~\cite{wei2023towards}   & 13.2 &  23.4 \\
         ~~w/ ACR~\cite{wei2023towards} & 13.8 & 23.3  \\
        \cmidrule(r){1-1} \cmidrule(l){2-3}
\rowcolor{mygray}
         ~~w/ \name & \textbf{19.7} & \textbf{25.0} \\

        \bottomrule

    \end{tabular}
}
}
\end{table}

\subsection{Results}\label{sec:main_results}
We first conduct experiments on the four representative benchmark datasets with different class imbalance ratios.
We denote the class imbalance ratio of labeled, unlabeled, and test data as $\gamma_l$, $\gamma_u$, and $\gamma_t$, respectively.
Our method is compared with five competitive baseline approaches, i.e., FixMatch~\cite{Sohn2020}, CReST+~\cite{Wei2021}, DASO~\cite{Oh2022}, ACR~\cite{wei2023towards}, and CPE~\cite{Ma2023}. Note that for a fair comparison, we first compare with ACR in the ReaLTSSL setting, where the unlabeled class distribution is unknown and inaccessible. Specifically, we compare our vanilla \name with ACR's variant that removes its pre-defined anchor distributions, denoted as ACR$^\dagger$. Then we implement SimPro$^\star$ by also alleviating the anchor distributions in our \name framework, comparing SimPro$^\star$ with the original ACR and CPE.

\paragraph{Main results and comparison with SOTA baselines.}  The results are presented in \cref{tab:main_cifar10} (for CIFAR10-LT), \cref{tab:main_cifar100} (for CIFAR100-LT and STL10-LT), and \cref{tab:main_imagenet} (for ImageNet-127/1k).
It can be concluded that our method consistently outperforms the competitors across all distributions of unlabeled data and achieves SOTA performance.
Notably, \name exhibits significant performance improvements on our two newly introduced distributions of unlabeled data: middle and head-tail. This substantiates the robust generalization capabilities of \name across various distributions that could potentially appear in real-world scenarios.

It is worth noting that compared to CIFAR10/100-LT,
STL10-LT is a more challenging dataset that mirrors the real-world data distribution scenarios:
an \emph{unknown} distribution for the unlabeled data.
The results in \cref{tab:main_cifar100} demonstrate the significant improvements of \name over baseline methods.

Moreover, we also conduct experiments on ImageNet-127, whose test dataset is imbalanced and consistent with the labeled data and unlabeled data.
However, this is not suitable as a benchmark for long-tail learning,
as biased classifiers tend to perform well in such scenarios, which is precisely what we aim to avoid.
Therefore, we resample it to achieve a uniform test distribution ($\gamma_t=1$).
The results highlight that our \name achieves substantial performance enhancements when evaluated against this balanced test dataset.
Beyond this, we further conduct experiments on ImageNet-1k to validate the performance of our method across a broader range of classes.
The results in \cref{tab:main_imagenet} demonstrate that our \name achieves state-of-the-art performance on ImageNet-1k.

\begin{table}
\caption{The impact of the predefined anchor distribution in ACR and CPE~\cite{Ma2023} on CIFAR10-LT with $\gamma_l=150, N_1=500$, and $M_1=4000$.
    $\star$ denotes that we use the predefined anchor distributions to estimate $P(y|\bm{\pi})$ in our \name.
    See more analysis in the main text and more results in \cref{sec:more_exp}.
}
\label{tab:anchor}

\centering
 \resizebox{1.0\linewidth}{!}{%
    \begin{tabular}{lccccc}
         \toprule
            & \multicolumn{1}{c}{$\gamma_u=150$} & \multicolumn{1}{c}{$\gamma_u=1$}  & \multicolumn{1}{c}{$\gamma_u=1/150$} & \multicolumn{1}{c}{$\gamma_u=150$} & \multicolumn{1}{c}{$\gamma_u=150$} \\
        \cmidrule(lr){2-4} \cmidrule(l){5-6}
                                    & \multicolumn{1}{c}{consistent} & \multicolumn{1}{c}{uniform}  & \multicolumn{1}{c}{reversed} & \multicolumn{1}{c}{middle} & \multicolumn{1}{c}{head-tail} \\
        \cmidrule{1-1} \cmidrule(lr){2-4} \cmidrule(l){5-6}
        CPE & 76.8 & 81.0 & 80.8 & -- & -- \\
        ACR & {77.0} & 91.3 & 81.8 & 77.9    & 79.0 \\
        \cmidrule{1-1} \cmidrule(lr){2-4} \cmidrule(l){5-6}
        \name & 74.2 & {93.6} & {83.5} & \textbf{82.6} & \textbf{81.0} \\
        SimPro$^\star$ & \textbf{80.0} & \textbf{94.1} & \textbf{85.0} & -- & -- \\
        \bottomrule
    \end{tabular}
}
\end{table}

\paragraph{The results of SimPro$^{\star}$ using anchor distributions.} To investigate the impact of the anchor distributions in ACR and CPE~\cite{Ma2023}, we also incorporate them into our approach, referred to as SimPro$^\star$.
Instead of calculating the distribution distance and adjusting the consistency regularization as in ACR or employing multiple anchor experts as in CPE,
our usage of these anchors is more straightforward:
after training for five epochs, we calculate the distance between our estimated distribution $P(y|\bm{\pi})$ and the three anchor distributions (i.e. consistent, uniform, and reversed).
This calculation helps us predict the actual distribution and construct the Bayes classifier.
Then we fix the marginal distribution parameters $\bm{\pi}$ in the remainder of the training.

The results in \cref{tab:anchor} indicate that
(1) the usage of anchor distributions can significantly enhance the performance of SimPro$^{\star}$, consistently outperforming the original ACR and CPE;
(2) our estimation for $\bm{\pi}$ is accurate (\cref{fig:KL} further validates the accurate estimation of $\bm{\pi}$);
(3) when the pre-defined anchors fail to cover the evaluated distributions (middle and head-tail), \name outperforms ACR by a large margin;
(4) even compared to the original ACR, \name exhibits enhanced performance across all scenarios except the consistent distribution.
This demonstrates the superior \emph{robustness} and \emph{generalization ability} of our method. We believe these advantages guarantee a better \emph{adaptable} nature for \name in real applications and are more valuable than the accuracy improvements when using the anchor distributions.

\begin{figure}[t]
    \centering
    \includegraphics[width=0.9\linewidth]{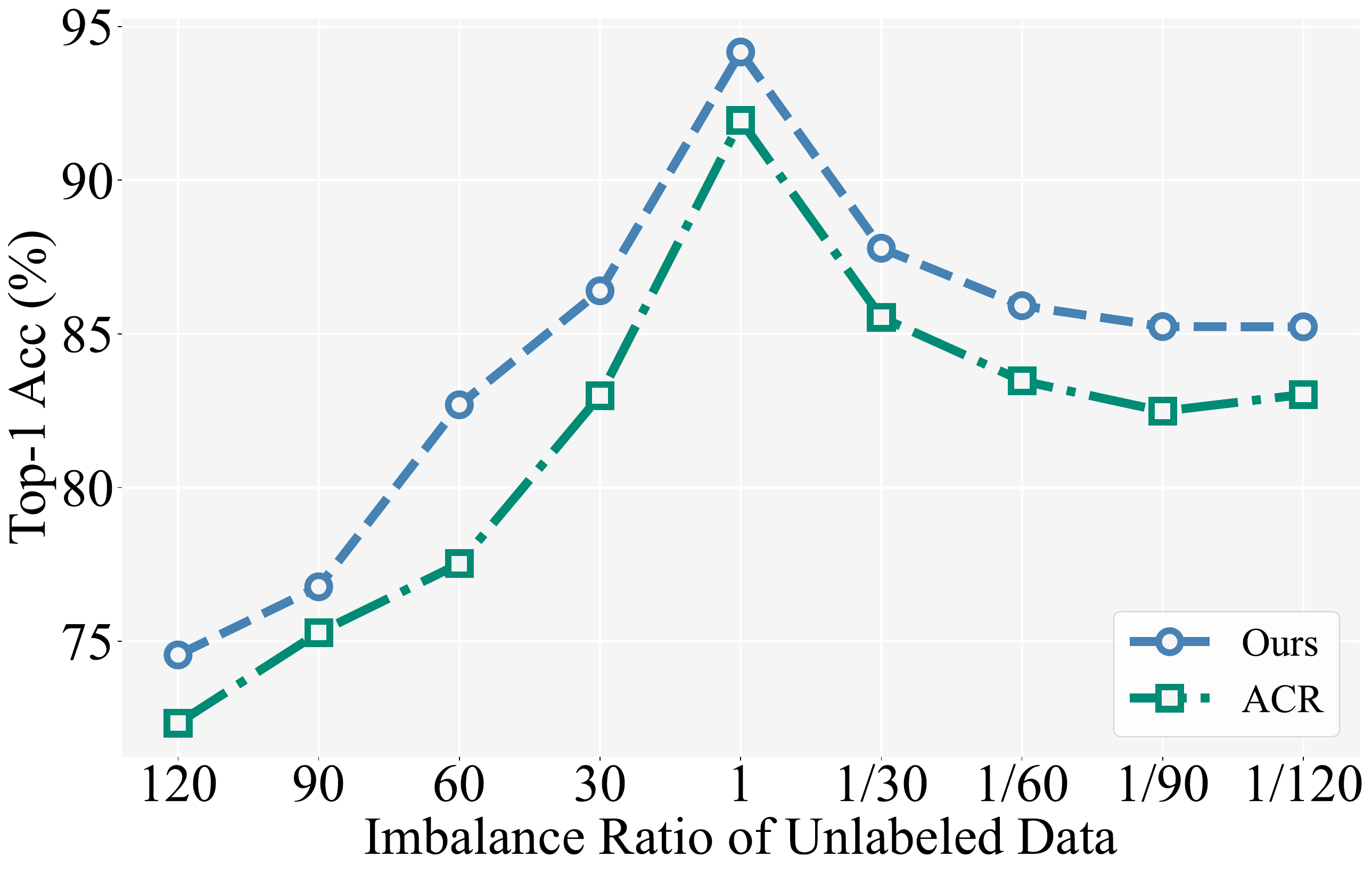}
    \caption{
        Test the performance under more imbalance ratios on CIFAR10-LT with $\gamma_l=150$, $N_1 = 500$, and  $M_1 = 4000$.
    }
     \label{fig:imb}
\end{figure}

\paragraph{Evaluation under more imbalance ratios.}  \cref{fig:imb} reports the performance of \name under more imbalance ratios of unlabeled data.
The results indicate that our method consistently outperforms ACR across all imbalance ratios, further substantiating the robustness of our method.

\begin{table}
\caption{
    Ablation study of estimating the marginal distribution $P(y|\bm{\pi})$ in M-step (\cref{eq:pi}) and using it for constructing Bayes classifier in E-step (\cref{eq:log_likelihood}).
    We conduct the experiments on CIFAR10-LT with $\gamma_l=150, N_1=500$, and $M_1=4000$.
}
\label{tab:estimation}

\centering
 \resizebox{1.0\linewidth}{!}{%
    \begin{tabular}{ccccccc}
         \toprule
        \multicolumn{2}{c}{Distribution Estimation} & \multicolumn{1}{c}{$\gamma_u=150$} & \multicolumn{1}{c}{$\gamma_u=1$}  & \multicolumn{1}{c}{$\gamma_u=1/150$} & \multicolumn{1}{c}{$\gamma_u=150$} & \multicolumn{1}{c}{$\gamma_u=150$} \\
           \cmidrule(r){1-2} \cmidrule(l){3-7}
                     E-step                 & M step              & \multicolumn{1}{c}{consistent} & \multicolumn{1}{c}{uniform}  & \multicolumn{1}{c}{reversed} & \multicolumn{1}{c}{middle} & \multicolumn{1}{c}{head-tail} \\
           \cmidrule(r){1-2} \cmidrule(l){3-7}
        
          \xmark  & \xmark & {40.7} & {35.3} & {43.2} & {27.1} & {47.7} \\
          \xmark  & \cmark & {64.1} & {92.6} & {78.6} & {64.9} & {74.8} \\

           \cmidrule(r){1-2} \cmidrule(l){3-7}

\rowcolor{mygray}
          \cmark  & \cmark  & \textbf{74.2} & \textbf{93.6} & \textbf{83.5} & \textbf{82.6} & \textbf{81.0} \\
        \bottomrule
    \end{tabular}
}
\end{table}

\subsection{Analysis}\label{sec:analysis}

\paragraph{Ablation Study.}

\begin{table}
\vskip -0.1in
\caption{
Ablation study of $\mu\!=\!M/N$ (\cref{eq:l_u}) on CIFAR10-LT with $\gamma_l\!=\!150, N_1\!=\!500$, and $M_1\!=\!4000$.
For the baseline methods without our Bayes classifier, the performance drops significantly when we set $\mu\!=\!M/N\!=\!8$. This large $\mu$ leads to an imbalance between labeled and unlabeled samples in each mini-batch. In contrast, our \name is not affected by such imbalance thanks to the Bayes classifier (\cref{prop:phi}). Moreover, we effectively leverage the large number of unlabeled data for a more accurate estimate of the marginal distribution parameters $\bm{\pi}$.}
\label{tab:recipe}
\centering
 \resizebox{1.0\linewidth}{!}{%
    \begin{tabular}{lccccc}
         \toprule
         & & \multicolumn{1}{c}{$\gamma_u=150$} &  \multicolumn{1}{c}{$\gamma_u=1/150$} & \multicolumn{1}{c}{$\gamma_u=150$} & \multicolumn{1}{c}{$\gamma_u=150$} \\
        \cmidrule(l){3-6}
         &  $\mu\!=\!M/N$           & \multicolumn{1}{c}{consistent} &  \multicolumn{1}{c}{reversed} & \multicolumn{1}{c}{middle} & \multicolumn{1}{c}{head-tail} \\
        \cmidrule(r){1-2} \cmidrule(l){3-6}
          \multirow{2}{*}{FixMatch} & \xmark   & {62.9} & 59.9 & 64.3 &  58.3  \\
                                                            & \cmark   & {40.7} & {43.2} & {27.1} & {47.7} \\
        \cmidrule(r){1-2} \cmidrule(l){3-6}
          \multirow{2}{*}{ ~~w/ ACR  }  & \xmark   & {70.9} &  83.2 & 73.8 &  77.6 \\
          & \cmark   & {68.7} &   58.9 & 69.7    & 72.4 \\
        \cmidrule(r){1-2} \cmidrule(l){3-6}
          \multirow{2}{*}{ ~~w/ \name} & \xmark  & 52.7 & {78.8} & {58.8} & {71.5} \\
              &     \cellcolor{mygray}\cmark  & \cellcolor{mygray}\textbf{75.2} & \cellcolor{mygray}\textbf{83.5} & \cellcolor{mygray}\textbf{82.6} & \cellcolor{mygray}\textbf{81.0} \\

        \bottomrule
    \end{tabular}
    }
\end{table}

\begin{table}
\vskip -0.2in
\caption{
    Abalation study of $\alpha$ for balancing loss (\cref{eq:overall_loss}).
    The results indicate that $\alpha$ substantially improves the model's performance and prevents the model from overfitting to labeled data.
}
\label{tab:alpha}

\centering
 \resizebox{1.0\linewidth}{!}{%
    \begin{tabular}{cccccc}
         \toprule
       & \multicolumn{2}{c}{CIFAR10-LT ($\gamma_u=1$)} & \multicolumn{1}{c}{CIFAR100-LT ($\gamma_u=1$)} & \multicolumn{2}{c}{STL10-LT ($\gamma_u=$ N/A)}  \\
    \cmidrule(l){2-6}
        & \multicolumn{2}{c}{$N_1=500, M_1=4000$} & \multicolumn{1}{c}{$N_1=50,M_1=400$} & \multicolumn{2}{c}{$N_1=450,M=1\!\times\!10^5$}  \\
        \cmidrule(r){1-1}    \cmidrule(lr){2-3} \cmidrule(lr){3-3} \cmidrule(lr){4-4} \cmidrule(l){5-6}
       $\alpha$  & \multicolumn{1}{c}{$\gamma_l=150$} & \multicolumn{1}{c}{$\gamma_l=100$}  & \multicolumn{1}{c}{$\gamma_l=20$} & \multicolumn{1}{c}{$\gamma_l=20$} & \multicolumn{1}{c}{$\gamma_l=10$} \\
        \cmidrule(r){1-1} \cmidrule(lr){2-3} \cmidrule(lr){4-4}  \cmidrule(l){5-6} 
        
          \xmark  & 92.1 & 91.2 & 49.4 & 76.4 & 80.0\\
        \cmidrule(r){1-1} \cmidrule(lr){2-3} \cmidrule(lr){4-4}  \cmidrule(l){5-6} 
\rowcolor{mygray}
          \cmark  & \textbf{93.6} & \textbf{93.8} & \textbf{52.2} & \textbf{83.0} & \textbf{85.2} \\
        \bottomrule
    \end{tabular}
}
\end{table}

\begin{figure*}[t]
    \centering
    \includegraphics[width=0.9\linewidth]{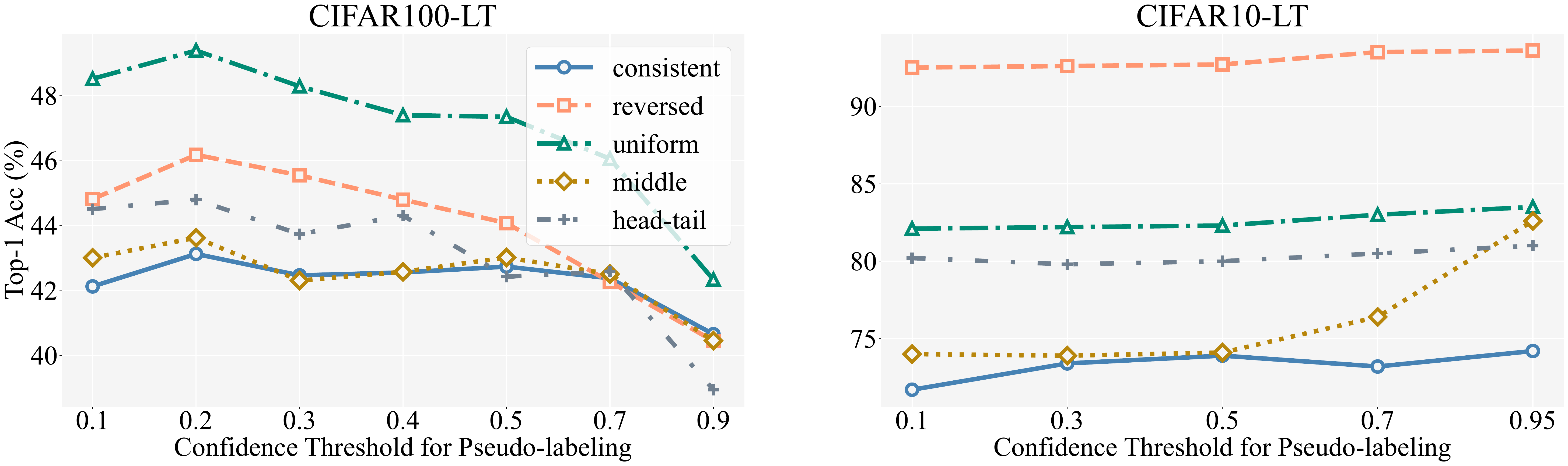}
    \caption{
        Sensitive analysis of the confidence threshold $t$ on CIFAR100-LT with $\gamma_l=20$, $N_1\!=\!50$, and $M_1\!=\!400$ and CIFAR10-LT with $\gamma_l\!=\!150$, $N_1\!=\!500$, and $M_1\!=\!4000$. 
    The optimal performance is consistently achieved  across different settings when the threshold is set at $t=0.2$ and $0.95$, respectively.
    }
     \label{fig:threshold}
 \end{figure*}

\begin{figure*}[t]
    \centering
    \includegraphics[width=1.0\linewidth]{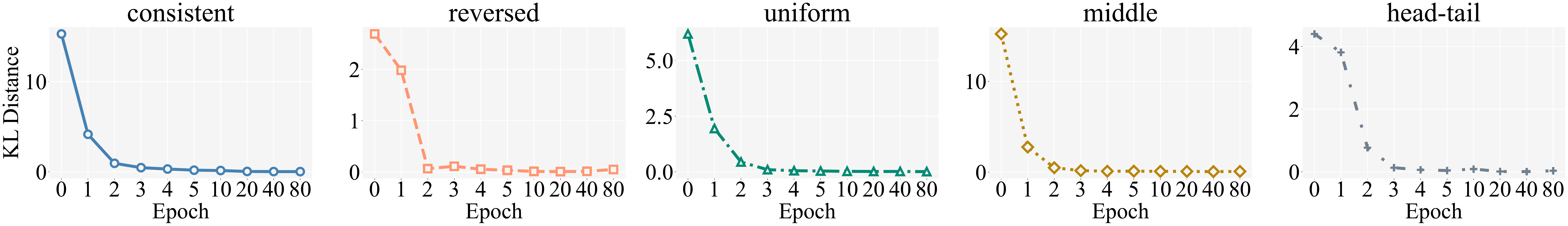}
    \vskip -0.05in
    \caption{
        Visualization of the quality of the estimated distribution on CIFAR10-LT with $\gamma_l\!=\!150$, $N_1\!=\!500$, and $M_1\!=\!4000$.
        The KL distances reduce to near-zero values after very few epochs.
    }
    \label{fig:KL}
\end{figure*}

We conduct a series of ablation studies to validate the effectiveness of different components in \name.

\noindent (a) \emph{Marginal distribution estimation}.
We first investigate the impact of the estimation for $P(y|\bm{\pi})$ (M-step, \cref{prop:pi}) and its usage in building the Bayes classifier for pseudo-label generation (E-step, \cref{prop:phi}).
The results in \cref{tab:estimation} substantiate the high effectiveness and the necessity of this estimation in driving the success of our approach,
thereby validating our methodology both theoretically and practically.

\noindent (b) \emph{More unlabeled samples (larger $\mu$) in each iteration.}
As mentioned in the experimental setup, we manually set the ratio between unlabeled and labeled samples in each training iteration as $\mu\!=\!M/N$ (\cref{eq:l_u}).
In ACR or Fixmatch, this ratio is set as $2$.
To investigate the impact of this adjustment, we adopt our setting of $\mu\!=\!M/N\!=\!8$ for ACR and Fixmatch.
The results in \cref{tab:recipe} demonstrate that our method can effectively leverage the unlabeled data for a more accurate estimation of the marginal distribution parameters $\bm{\pi}$.
However, the baseline methods suffer from an imbalanced number of labeled/unlabeled samples, because of the absence of our Bayes classifier derived in \cref{prop:phi}.

\noindent (b) \emph{Scaling factor $\alpha$}. 
As elucidated in \cref{sec:implementation}, we introduce a scaling factor $\alpha\!=\!\mu \cdot N/M$ (\cref{eq:overall_loss}) to mitigate the risk of overfitting.
This measure is primarily due to memory and training environment limitations, which restrict the feasible batch size and $\mu$,  when $M \gg N$.
In the test configurations detailed in \cref{tab:alpha}, the ratio $M/N$ is about $30$ for CIFAR and $67$ for STL10-LT.
An insufficient $\mu$ results in a disproportionately high number of labeled data within a mini-batch, potentially leading to overfitting.
Hence, we incorporate the $\alpha$ to balance the losses between labeled and unlabeled data.
Our empirical findings demonstrate that the use of this simple parameter $\alpha$ can significantly enhance model performance, particularly STL10-LT, where there is a substantial disparity between the sizes of labeled and unlabeled data.

\paragraph{Hyperparameter Sensitivity.}

    As outlined in \cref{sec:setup}, we discover that reducing the threshold value improves performance for the CIFAR100-LT dataset.
    The rationale behind adjusting the confidence threshold is based on the observation that an increase in the number of classes typically results in a corresponding decrease in the confidence of the predictions.
    Consequently, it becomes necessary to lower the threshold to accommodate this change in confidence levels.
    Sensitivity analysis regarding the threshold value is presented in \cref{fig:threshold}.
    It is consistently observed across different settings that the optimal performance is achieved when the threshold is set at $t=0.2$ and $t=0.95$ for the CIFAR100-LT and CIFAR10-LT, respectively.
    Moreover, to compare with ACR, we also conduct a sensitivity analysis of the confidence threshold $t$ for ACR.
    The results in \cref{fig:threshold_ACR} of \cref{sec:more_exp} demonstrate that lowering the threshold does not improve performance for ACR.

\paragraph{Visualization of Estimation Quality.}

        Our study includes a visualization of the estimated distribution quality in \cref{fig:KL}.
        The vertical axis quantifies the Kullback-Leibler (KL) divergence, which measures the deviation of the estimated distribution from the ground truth.
        The results indicate a significant improvement in estimation accuracy after only a few training epochs.
        This empirical evidence validates the effectiveness of the theoretically derived estimation method of distribution, as outlined in \cref{prop:pi}.

\section{Conclusion}
\label{sec:conclusion}

In this paper, we introduce \name, a novel probabilistic framework for realistic long-tailed semi-supervised learning (ReaLTSSL).
This framework represents a pioneering advancement in the field by innovatively enhancing the Expectation-Maximization (EM) algorithm. 
The key innovation lies in the explicit separation of the estimation process for conditional and marginal distributions.
In the M-step, this separation allows for the derivation of a \emph{closed-form} solution for the marginal distribution parameters.
Additionally, \name optimizes the parameters of the conditional distribution via gradient descent, facilitating the learning of a \emph{Bayes classifier}.
In the E-step. the Bayes classifier, in turn, generates high-quality pseudo-labels.
\name is characterized by its \emph{elegant} theoretical underpinnings and its \emph{ease} of implementation, which requires only minimal modifications to existing codebases.
Furthermore, we incorporate two innovative class distributions specifically for unlabeled data.
These distributions provide a more comprehensive and realistic evaluation of ReaLTSSL algorithms.
Empirical evidence from various benchmarks demonstrates that \name consistently delivers state-of-the-art performance, highlighting its robustness and superior generalization capabilities.

\section*{Acknowledgements}

%\textbf{Do not} include acknowledgements in the initial version of
%the paper submitted for blind review.
%
%If a paper is accepted, the final camera-ready version can (and
%usually should) include acknowledgements.  Such acknowledgements
%should be placed at the end of the section, in an unnumbered section
%that does not count towards the paper page limit. Typically, this will 
%include thanks to reviewers who gave useful comments, to colleagues 
%who contributed to the ideas, and to funding agencies and corporate 
%sponsors that provided financial support.

This work was supported in part by the National Key R\&D Program of China under Grant 2021ZD0140407, in part by the National Natural
Science Foundation of China under Grants 62276150, 62321005 and 42327901.

\section*{Impact Statement}

This paper presents work whose goal is to advance the field of Machine Learning. There are many potential societal consequences of our work, none which we feel must be specifically highlighted here.

% In the unusual situation where you want a paper to appear in the
% references without citing it in the main text, use \nocite
%\nocite{langley00}

\bibliography{main}
\bibliographystyle{icml2024}

%%%%%%%%%%%%%%%%%%%%%%%%%%%%%%%%%%%%%%%%%%%%%%%%%%%%%%%%%%%%%%%%%%%%%%%%%%%%%%%
%%%%%%%%%%%%%%%%%%%%%%%%%%%%%%%%%%%%%%%%%%%%%%%%%%%%%%%%%%%%%%%%%%%%%%%%%%%%%%%
% APPENDIX
%%%%%%%%%%%%%%%%%%%%%%%%%%%%%%%%%%%%%%%%%%%%%%%%%%%%%%%%%%%%%%%%%%%%%%%%%%%%%%%
%%%%%%%%%%%%%%%%%%%%%%%%%%%%%%%%%%%%%%%%%%%%%%%%%%%%%%%%%%%%%%%%%%%%%%%%%%%%%%%
\newpage
\appendix
\onecolumn

\section{Details of the Probabilistic Model}
\label{sec:detailed_model}

We provide a detailed derivation and analysis to demonstrate that the probabilistic model is correctly defined with an explicit parameterization of $p(x;\xi)$.
Given the independence of parameters through conditional parameterization~\cite{Koller2009}, we can decompose the joint probability distribution as follows:
\begin{equation}
    p(x,y;\theta,\pi,\xi) = p(x;\xi)p(y|x;\theta,\pi) = p(y;\pi)p(x|y;\theta,\xi).
\end{equation}
Applying Bayes' rule, we obtain:
\begin{equation}
    \frac{p(y|x;\theta,\pi)}{p(y;\pi)} = \frac{p(x|y;\theta,\xi)}{p(x;\xi)}.
\end{equation}
It is evident that $\pi$ and $\xi$ appear only on the left and right sides of the equation, respectively, indicating that the equation is neither a function of $\pi$ nor $\xi$ but is parameterized solely by $\theta$.
We define the above equation as $g(x,y;\theta)$, that is:
\begin{equation}
    \frac{p(x|y;\theta,\xi)}{p(x;\xi)} = \frac{p(y|x;\theta,\pi)}{p(y;\pi)} = g(x,y;\theta).
\end{equation}
Returning to the equation in the main paper, we have: 
\begin{equation}
    p(y|x;\theta,\pi) = \frac{p(y;\pi)p(x|y;\theta,\xi)}{p(x;\xi)} = p(y;\pi)g(x,y;\theta).
\end{equation}
Although we explicitly parameterize $p(x;\xi)$, it is clear that $p(y|x;\theta,\pi)$ is parameterized solely by $\theta$ and $\pi$, and is independent of $\xi$.
In fact, the fitting target of the network parameters $\theta$ is $g(x,y;\theta) = p(x|y;\theta,\xi)/p(x;\xi)$.

Since we did not explicitly parameterize $p(x)$ in our framework, when $x$ is not a condition, the parameters of the relevant notions omit the parameters of $p(x)$, such as $p(x)$, $p(x|y)$, $p(x,y)$, etc.
This may lead to a potential misunderstanding, as:
\begin{equation}
    p(x) = \sum_{y}p(x|y;\theta)p(y;\pi)
\end{equation}
suggests that $p(x)$ seems to be parameterized by $\theta$ and $\pi$.
However, if we recover the omitted parameter of $p(x)$, we have: 
\begin{equation}
    p(x;\xi) = \sum_{y}p(x|y;\theta,\xi)p(y;\pi) = p(x;\xi) \sum_{y} g(x,y;\theta) p(y;\pi) = p(x;\xi) \sum_{y} p(y|x;\theta,\pi) = p(x;\xi),
\end{equation}
which is consistent with the explicit parameterization of $p(x;\xi)$.

Therefore, the probabilistic model is correctly defined without the explicit parameterization of $p(x)$.

\section{Experimental Setup}
\label{sec:setup}

\paragraph{Datasets.}

    To validate the effectiveness of \name, we employ five commonly used SSL datasets,
    CIFAR10/100~\cite{krizhevsky2009learning}, STL10~\cite{pmlr-v15-coates11a}, ImageNet-127~\cite{Fan2022} and orginal ImageNet-1k~\cite{Deng2009}.
    Following the methodology described in ACR~\cite{wei2023towards}, we denote the number of samples for each category in the labeled, unlabeled dataset as $N_1\!\ge\!\cdots\!\ge\!N_K$, $M_1\!\ge\!\cdots\!\ge\!M_K$, respectively, where $1,\cdots,K$ are the class indices.
    We define $\gamma_l$, $\gamma_u$, and $\gamma_t$ as the class imbalance ratios for labeled, unlabeled, and test data, respectively.
    We specify ‘LT’ for those imbalanced variants.
    These ratios are calculated as $\gamma_l\!=\!N_1/N_K$ and $\gamma_u\!=\!M_1/M_K$.
    The sample number of the $k$-th class follows an exponential distribution, expressed as $N_k\!=\!N_1 \cdot \gamma_l^{-\frac{k-1}{K-1}}$ for labeled and $M_k\!=\!M_1 \cdot \gamma_u^{-\frac{k-1}{K-1}}$ for unlabeled data.

    As illustrated in \cref{fig1:distribution}, for the CIFAR10/100 datasets, we constructed five class distributions to test the performance of different algorithms under more general settings.
    Regarding the STL10 dataset, due to the unknown ground-truth labels of the unlabeled data, we approach the experiments by controlling the imbalance ratio of the labeled data.

    For ImageNet-127, we follow the original setting in \citet{Fan2022} ($\gamma_l\!=\!\gamma_u\!=\!\gamma_t\approx286$).
    %However, this is not suitable as a benchmark for long-tail learning, as biased classifiers tend to perform well in such scenarios,
    %which is precisely what we aim to avoid.
    %Therefore, we resample the test dataset to achieve a uniform distribution ($\gamma_t\!=\!1$).
    Nevertheless, this approach does not serve as an appropriate benchmark for long-tail learning.
    In these scenarios, biased classifiers often exhibit high performance, which is exactly the outcome we seek to prevent.
    Consequently, we also resample the test dataset to ensure a uniform distribution ($\gamma_t\!=\!1$).
    Following \citet{Fan2022}, we reduce the image resolution to $32 \times 32$ and $64 \times 64$ in response to resource constraints.

\paragraph{Training hyper-parameters.}

    Our experimental setup mainly follows FixMatch~\cite{Sohn2020} and ACR~\cite{wei2023towards}.
    For example, we employ the Wide ResNet-28-2~\cite{zagoruyko2016wide} on CIFAR10/100 and STL10, and ResNet-50~\cite{He2016} on ImageNet-127.
    All models are optimized with SGD.
    Several settings are slightly different from those in ACR: as outlined in \cref{sec:implementation}, to achieve a balanced training,
    we set the ratio between unlabeled and labeled samples in each batch as $\mu\!=\!M/N$ ($8$ on CIFAR, $16$ on STL10-LT, and $2$ on ImageNet-127, 
    ),
    where $M, N$ are the total sample numbers of unlabeled/labeled data.
    In contrast, $\mu$ is set as $2$ for all datasets in ACR. The effectiveness of this adjustment is validated in \cref{tab:recipe}.
    
    The batch size for labeled data is $64$ on CIFAR10/100 and STL10-LT, and $32$ on ImageNet-127.
    To ensure a fair comparison, the training epochs are reduced to $86$ on CIFAR10/100 and STL10-LT, and $500$ on ImageNet-127.
    The initial learning rate $\eta$ is linearly scaled to $0.17$ on CIFAR10/100 and STL10-LT, and $0.01$ on ImageNet-127, which decays with a cosine schedule~\cite{loshchilov2016sgdr} as in ACR.
    
    Regarding the hyperparameter $\tau$ used in \cref{eq:l_l}, we follow the guidelines from~\citet{Menon2021} and set $\tau\!=\!2.0$ and $1.0$ for CIFAR10-LT/STL10-LT and CIFAR100-LT/ImageNet-127, respectively.

    For the confidence threshold $t$ in \cref{eq:FixMatch}), we set $t\!=\!0.95$ on CIFAR10-LT/STL10-LT following \citet{Sohn2020}. We adjust it to $0.2$ on CIFAR100-LT/Imagenet-127, as we observe that reducing the threshold enhances performance (\cref{fig:threshold}).
    
    Specifically, the settings on ImageNet-1k are identical to those on ImageNet-127.

\section{More Experimental Results}
\label{sec:more_exp}

\begin{table}[H]
  \centering
  \begin{minipage}{0.48\textwidth}
    \caption{The impact of the predefined anchor distribution in ACR~\cite{wei2023towards} on CIFAR10-LT with $\gamma_l=100, N_1=500$, and $M_1=4000$.
    $\star$ denotes that we use the predefined anchor distributions to estimate $P(y|\bm{\pi})$ in our \name.
    }
    \label{tab:anchor_cifar10_gamma100}

    \centering
     \resizebox{1.0\linewidth}{!}{%
        \begin{tabular}{lccccc}
           \toprule
           & \multicolumn{1}{c}{$\gamma_u=100$} & \multicolumn{1}{c}{$\gamma_u=1$}  & \multicolumn{1}{c}{$\gamma_u=1/100$} & \multicolumn{1}{c}{$\gamma_u=100$} & \multicolumn{1}{c}{$\gamma_u=100$} \\
            \cmidrule(lr){2-4} \cmidrule(l){5-6}
            & \multicolumn{1}{c}{consistent} & \multicolumn{1}{c}{uniform}  & \multicolumn{1}{c}{reversed} & \multicolumn{1}{c}{middle} & \multicolumn{1}{c}{head-tail} \\
            \cmidrule{1-1} \cmidrule(lr){2-4} \cmidrule(l){5-6}
            ACR & {81.6} & 92.1 & 85.0 & 73.6    & 79.8 \\
            \cmidrule{1-1} \cmidrule(lr){2-4} \cmidrule(l){5-6}
            \name & 80.7 & {93.8} & {85.8} & \textbf{84.8} & \textbf{83.0} \\
            SimPro$^\star$ & \textbf{82.7} & \textbf{94.3} & \textbf{86.0} & -- & -- \\
            \bottomrule
        \end{tabular}
    }
      \end{minipage}
      \hspace{0.5ex}
  \begin{minipage}{0.48\textwidth}
    \centering
\caption{The impact of the predefined anchor distribution in ACR~\cite{wei2023towards} on CIFAR100-LT with $\gamma_l=20, N_1=50$, and $M_1=400$.
    $\star$ denotes that we use the predefined anchor distributions to estimate $P(y|\bm{\pi})$ in our \name.
}
\label{tab:anchor_cifar100}

\centering
 \resizebox{1.0\linewidth}{!}{%
    \begin{tabular}{lccccc}
       \toprule
       & \multicolumn{1}{c}{$\gamma_u=20$} & \multicolumn{1}{c}{$\gamma_u=1$}  & \multicolumn{1}{c}{$\gamma_u=1/20$} & \multicolumn{1}{c}{$\gamma_u=20$} & \multicolumn{1}{c}{$\gamma_u=20$} \\
        \cmidrule(lr){2-4} \cmidrule(l){5-6}
        & \multicolumn{1}{c}{consistent} & \multicolumn{1}{c}{uniform}  & \multicolumn{1}{c}{reversed} & \multicolumn{1}{c}{middle} & \multicolumn{1}{c}{head-tail} \\
        \cmidrule{1-1} \cmidrule(lr){2-4} \cmidrule(l){5-6}
        ACR & {44.9} & 52.2 & 42.3 & 42.6 & 42.6 \\
        \cmidrule{1-1} \cmidrule(lr){2-4} \cmidrule(l){5-6}
        \name & 43.1 & {52.3} & {45.5} & \textbf{43.6} & \textbf{44.8} \\
        SimPro$^\star$ & \textbf{45.9} & \textbf{53.8} & \textbf{46.0} & -- & -- \\
        \bottomrule
    \end{tabular}
}

  \end{minipage}
\end{table}

\begin{figure}[H]
    \centering
    \includegraphics[width=0.8\linewidth]{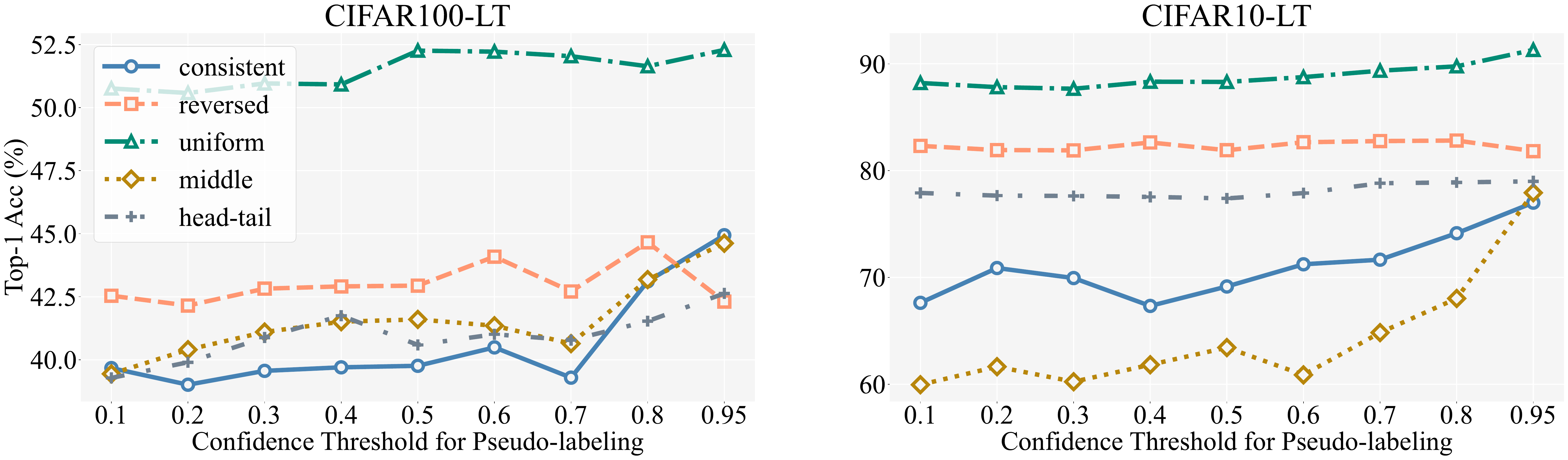}
    \caption{
        Sensitive analysis  of the confidence threshold $t$ for ACR~\cite{wei2023towards}  on CIFAR100-LT with $\gamma_l=20$, $N_1\!=\!50$, and $M_1\!=\!400$ and CIFAR10-LT with $\gamma_l\!=\!150$, $N_1\!=\!500$, and $M_1\!=\!4000$. 
    The optimal performance is achieved  across different settings when the threshold is set at $0.95$.
    }
    \label{fig:threshold_ACR}
 \end{figure}

\newpage

\section{Proof of \Cref{prop:pi}}
\label{sec:proof_pi}

\begin{proof}
    We employ the method of Lagrange multipliers to find the optimal values of $\bm{\hat\pi}_l$ and $\bm{\hat\pi}_u$ that maximize the $\mathcal{Q}$ function subject to the constraints of probability distributions (i.e., the elements of $\bm{\pi}_l$ and $\bm{\pi}_u$ must sum to 1).
    Let $\lambda_l$ and $\lambda_u$ be the Lagrange multipliers for these constraints.
    The Lagrangian $\mathcal{L}$ can be formulated as \footnote{$y$ is the one-hot label, but when $y$ is the subscript of a variable, it represents the $y$-th category, that is,  $\phi_y = \phi_{\argmax y}$. we omit this difference without affecting the understanding.}:

\begin{align}
    \mathcal{L}(\bm{\pi}_l, \bm{\pi}_u, \lambda_l, \lambda_u) &= \sum_{i}\log P(y_i;\bm{\pi}_l) \nonumber + \sum_{j,y}P(y|x_j;\bm{\theta}',\bm{\pi}')\log P(y;\bm{\pi}_u) \nonumber \\
                                                              &- \lambda_l \left( \sum_{y} \pi_{ly} - 1 \right)  - \lambda_u \left( \sum_{y} \pi_{uy} - 1 \right).
\end{align}

The partial derivatives of $\mathcal{L}$ with respect to $\bm{\pi}_l$ and $\bm{\pi}_u$ are calculated as follows:

\begin{align}
\frac{\partial \mathcal{L}}{\partial \bm{\pi}_l} &= \frac{\partial}{\partial \bm{\pi}_l} \left( \sum_{i}\log P(y_i;\bm{\pi}_l) \right) - \lambda_l \frac{\partial}{\partial \bm{\pi}_l} \left( \sum_{y} \pi_{ly} - 1 \right), \\
\frac{\partial \mathcal{L}}{\partial \bm{\pi}_u} &= \frac{\partial}{\partial \bm{\pi}_u} \left( \sum_{j,y}P(y|x_j;\bm{\theta}',\bm{\pi}')\log P(y;\bm{\pi}_u) \right) - \lambda_u \frac{\partial}{\partial \bm{\pi}_u} \left( \sum_{y} \pi_{uy} - 1 \right).
\end{align}

Due to fact that $P(y;\bm{\pi})$ is a categorical distribution, the derivatives are:

\begin{align}
    \frac{\partial \mathcal{L}}{\partial \pi_{ly}} &= \sum_{i} \frac{\mathbf{1}_{\{y_i = y\}}}{\pi_{ly}} - \lambda_l, \\
\frac{\partial \mathcal{L}}{\partial \pi_{uy}} &= \sum_{j} P(y|x_j;\bm{\theta}',\bm{\pi}') \frac{1}{\pi_{uy}} - \lambda_u.
\end{align}

Setting these partial derivatives to zero and solving for $\pi_{ly}$ and $\pi_{uy}$:

\begin{align}
    \hat\pi_{ly} &= \frac{\sum_{i} \mathbf{1}_{\{y_i = y\}}}{\lambda_l}, \\
\hat\pi_{uy} &= \frac{\sum_{j} P(y|x_j;\bm{\theta}',\bm{\pi}')}{\lambda_u}.
\end{align}

Applying the constraint that the sum of probabilities equals 1, we get:

\begin{align}
    \sum_{y} \hat\pi_{ly} = 1 &\Rightarrow \lambda_l = \sum_{y}\sum_{i} 1 = N, \\
    \sum_{y} \hat\pi_{uy} = 1 &\Rightarrow \lambda_u = \sum_{y}\sum_{j} P(y|x_j;\bm{\theta}',\bm{\pi}') = M.
\end{align}

Therefore, the optimal solutions are:

\begin{equation}
\bm{\hat\pi}_l = \frac{1}{N}\sum_{i=1}^{N}{y_i}, \quad \bm{\hat\pi}_u = \frac{1}{M}\sum_{j=1}^{M}P(y|x_j;\bm{\theta}',\bm{\pi}').
\end{equation}

\end{proof}

\section{Proof of \Cref{prop:phi}}
\label{sec:proof_phi}

\begin{proof}

We structure our proof in two parts: First, we validate the proposition when training exclusively with labeled data. Then, we extend our analysis to include scenarios incorporating unlabeled data. This approach stems from our threshold-based strategy for filtering low-confidence pseudo labels in the training process. Initially, only labeled data are used, gradually integrating unlabeled data as training progresses.

\subsection*{Case 1: Labeled Data Only}

Our proof begins by revisiting the definition of $\mathcal{Q}(\bm{\theta},\bm{\pi};\bm{\theta}',\bm{\pi}')$ with respect to $\bm{\theta}$:
\begin{align}
    \mathcal{Q}(\bm{\theta}) &= \sum_{i} \log \frac{\exp(f_{\bm{\theta}}(x_i,y_i))}{\sum_{y'}\phi_{y'}\exp(f_{\bm{\theta}}(x,y'))} \nonumber \\
                             &= \sum_{i} \left( \log \frac{\phi_{y_i}\exp(f_{\bm{\theta}}(x_i,y_i))}{\sum_{y'}\phi_{y'}\exp(f_{\bm{\theta}}(x,y'))} - \log \phi_y \right).
\end{align}
Ignoring the constant term, maximizing $\mathcal{Q}(\bm{\theta})$ is equivalent to minimizing the empirical risk:
\begin{equation}
    R_{\text{emp}}(\bm{\theta}) = - \frac{1}{N} \sum_i \log \frac{\phi_{y_i}\exp(f_{\bm{\theta}}(x_i,y_i))}{\sum_{y'}\phi_{y'}\exp(f_{\bm{\theta}}(x,y'))}.
\end{equation}
The empirical risk serves as an approximation of the expected risk, with $\mathcal{D}_l$ denoting the distribution of labeled data:
\begin{align}
    R_{\text{exp}}(\bm{\theta}) &= - \mathbb{E}_{(x,y) \sim \mathcal{D}_l} \log Q(y|x) \nonumber \\
                                &= - \int_x \sum_y P(y|x)  \log Q(y|x) \, \dif x \nonumber \\
                                &= \int_x H(P(y|x)) + D_{\text{KL}}(P(y|x)||Q(y|x)) \, \dif x,
\end{align}
where $Q(y|x)$ is defined as:
\begin{equation}
Q(y|x) = \frac{\phi_{y}\exp(f_{\bm{\theta}}(x,y))}{\sum_{y'}\phi_{y'}\exp(f_{\bm{\theta}}(x,y'))}.
\end{equation}
With the non-negativity and zero-equality conditions of KL divergence, minimizing the expected risk implies:
\begin{equation}
    \frac{\phi_{y}\exp(f_{\bm{\theta}}(x,y))}{\sum_{y'}\phi_{y'}\exp(f_{\bm{\theta}}(x,y'))} = P(y|x) = \frac{\pi_{ly}P(x|y)}{P_l(x)}.
    \label{eq:optimal}
\end{equation}
We aim to validate the Bayes classifier:
\begin{equation}
    P(y|x;\bm{\theta},\bm{\pi}) = \frac{P(y;\bm{\pi})\exp(f_{\bm{\theta}}(x,y))}{\sum_{y'}P(y';\bm{\pi})\exp(f_{\bm{\theta}}(x,y'))},
\end{equation}
which leads to the formulation:
\begin{equation}
    \frac{\pi_y P(x|y)}{\hat P(x)} = \frac{\pi_y \exp(f_{\bm{\theta}}(x,y))}{\sum_{y'}\pi_{y'}\exp(f_{\bm{\theta}}(x,y'))}.
    \label{eq:Bayes}
\end{equation}
Integrating \cref{eq:optimal} with \cref{eq:Bayes} yields:
\begin{equation}
    \frac{\phi_y}{\sum_{y'}\phi_{y'} \exp(f_{\bm{\theta}}(x,y'))} = \frac{\pi_{ly} \hat P(x)}{P_l(x) \sum_{y'}\pi_{y'}\exp(f_{\bm{\theta}}(x,y'))}.
    \label{eq:phi_y}
\end{equation}
Summing over $y$ in \cref{eq:phi_y} leads to:
\begin{equation}
    \sum_{y} \phi_y = C = \frac{\hat P(x) \sum_{y'}\phi_{y'}\exp(f_{\bm{\theta}}(x,y'))}{P_l(x)\sum_{y'}\pi_{y'}\exp(f_{\bm{\theta}}(x,y'))]}.
    \label{eq:sum_phi}
\end{equation}
Substituting \cref{eq:sum_phi} into \cref{eq:phi_y} results in:
\begin{equation}
    \phi_y = C \pi_{ly}.
\end{equation}
As the constant $C$ becomes irrelevant in the logarithmic term of $\mathcal{Q}(\bm{\theta})$, in light of \cref{prop:pi}, we deduce the optimal $\bm{\hat\phi}$:
\begin{equation}
    \bm{\hat\phi} = \frac{1}{N}\sum_{i=1}^{N}{y_i}.
\end{equation}

\subsection*{Case 2: Labeled and Unlabeled Data}

Expanding our proof to include both labeled and unlabeled data, our optimization objective remains consistent: maximizing $\mathcal{Q}(\bm{\theta})$ by minimizing the empirical risk:
\begin{equation}
    R_{\text{emp}}(\bm{\theta}) = - \frac{1}{N+M} (\sum_i + \sum_{j,y}P(y|x_j;\bm{\theta}',\bm{\pi}')) \log Q(y|x).
\end{equation}
This risk approximates the expected risk over $\mathcal{D}$:
\begin{equation}
    R_{\text{exp}}(\bm{\theta}) = - \mathbb{E}_{(x,y) \sim \mathcal{D}} \log Q(y|x),
\end{equation}
where $\mathcal{D}$ represents the mixture distribution of both labeled and unlabeled data and its density is:
\begin{equation}
   P_\mathcal{D}(x,y) = \frac{N}{M+N} P_l(x,y) + \frac{M}{M+N} P_u(x) P(y|x;\bm{\theta}',\bm{\pi}').
\end{equation}
Acknowledging that $P(y|x;\bm{\theta}',\bm{\pi}')$ is a Bayes classifier, we conclude:
\begin{equation}
    P_\mathcal{D}(x,y) = \left ( \frac{N}{M+N} \pi_{ly} + \frac{M}{M+N} \pi_{uy} \right ) P(x|y),
\end{equation}
which leads to the formulation:
\begin{equation}
    P_\mathcal{D}(y) = \frac{N}{M+N} \pi_{ly} + \frac{M}{M+N} \pi_{uy}, \quad P_\mathcal{D}(x|y) = P(x|y).
\end{equation}
Following the same logic as in the labeled data case:
\begin{equation}
    \label{eq:fit}
    \frac{\phi_{y}\exp(f_{\bm{\theta}}(x,y))}{\sum_{y'}\phi_{y'}\exp(f_{\bm{\theta}}(x,y'))} = P_\mathcal{D}(y|x) = \frac{P_\mathcal{D}(y)P(x|y)}{P_\mathcal{D}(x)}.
\end{equation}
This results in:
\begin{equation}
    \phi_y = C \cdot P_\mathcal{D}(y).
\end{equation}
In accordance with \cref{prop:pi}, we determine the optimal $\bm{\hat\phi}$ as:

\begin{align}
    \bm{\hat\phi} &= \frac{N}{M+N} \bm{\hat\pi}_l + \frac{M}{M+N} \bm{\hat\pi}_u \nonumber \\
                &= \frac{1}{N+M}(\sum_i y_i + \sum_{j}P(y|x_j;\bm{\theta}',\bm{\pi}')).
\end{align}

\end{proof}

\section{Proof of \Cref{prop:regret}}
\label{sec:proof_regret}
\begin{proof}

We begin by examining $P(e;\bm{\hat\phi})$.
The decision error rate is mathematically defined as follows:
\begin{equation}
    \label{eq:def_decision_error}
    P(e) = \int_x P(e|x)P(x) \, \dif x, \quad
    P(e|x) = \begin{cases}
        P(-1|x) & \text{if decision is } +1; \\
        P(+1|x) & \text{if decision is } -1.
    \end{cases}
\end{equation}
This formulation quantifies the error in decision-making by integrating the conditional error rates across all possible outcomes.

Building upon the Bayes optimal classifier as outlined in \cref{eq:BOC_uniform}, the posterior probability essential for decision-making on the test set is expressed as:
\begin{equation}
    \label{eq:decision_posterior}
     P_d(y|x) \propto \exp(f_{\bm{\theta}}(x,y)).
\end{equation}
Here, $f_{\bm{\theta}}(x,y)$ represents the model's discriminative function, parameterized by $\bm{\theta}$, for decision outcome $y$ given an input $x$.

The ground truth distribution, as approximated by parameters $\bm{\hat\phi}$ and $\bm\theta$, is denoted by $P(y;\bm{\phi^*})$ and $P(x|y)$, respectively.
The formal relationship between the estimated and true distributions is expressed as follows:
\begin{equation}
    \label{eq:fit_truth}
    \frac{\hat\phi_{y}\exp(f_{\bm{\theta}}(x,y))}{\sum_{y'}\hat\phi_{y'}\exp(f_{\bm{\theta}}(x,y'))} = \frac{\phi^*_yP(x|y)}{\sum_{y'}\phi^*_{y'}P(x|y')},
\end{equation}
indicating a proportional relationship between the model's estimation and the true data distribution.

Integrating the formulation of posterior probability with the relationship between estimated and true distributions allows us to derive the following expression:
\begin{align}
     P_d(y|x) &\propto \exp(f_{\bm{\theta}}(x,y)) \nonumber \\
            &\propto \frac{\phi^*_y}{\hat\phi_y} P(x|y) \nonumber \\
        &\propto \frac{\phi^*_y}{\hat\phi_y} P(y|x),
\end{align}
where the final step is justified by the fact that the class distribution in the test set is uniform.

The decision criterion is thus formulated as:
\begin{equation}
    \label{eq:decision}
    \text{decision}=\begin{cases}
        +1 & \text{if } l(x) \ge \lambda; \\
        -1 & \text{if } l(x) \le \lambda,
    \end{cases}
\end{equation}
where 
\begin{equation}
    l(x) = \frac{P(+1|x)}{P(-1|x)}, \quad \lambda = \frac{\hat\phi_{+1}\phi^*_{-1}}{\hat\phi_{-1}\phi^*_{+1}}.
\end{equation}

We assume $\lambda \le 1$ without loss of generality, due to the symmetric nature of the decision problem.

The decision error rate is then expressed as:
\begin{equation}
    \label{eq:decision_error}
    P(e;\bm{\hat\phi}) = \int_x P(e|x;\bm{\hat\phi})P(x) \, \dif x, \quad P(e|x;\bm{\hat\phi}) = \begin{cases}
        P(-1|x) & \text{if } l(x) \ge \lambda; \\
        P(+1|x) & \text{if } l(x) \le \lambda.
    \end{cases}
\end{equation}
Minimizing the decision error rate is achieved when $\bm{\hat\phi} = \bm{\phi}^*$, leading to the Bayes decision error rate, which is the theoretical lower bound of error rates across all possible distributions.
\begin{equation}
    \label{eq:optimal_decision_error}
        \inf_{\bm\phi} P(e;\bm{\phi}) = P(e;\bm{\phi^*}) = \int_x P(e|x;\bm{\phi^*})P(x) \, \dif x, \quad P(e|x;\bm{\phi^*}) = \begin{cases}
        P(-1|x) & \text{if } l(x) \ge 1; \\
        P(+1|x) & \text{if } l(x) \le 1.
    \end{cases}
\end{equation}

Finally, comparing $P(e;\bm{\hat\phi})$ with the optimal decision error rate, we explore the difference:
\begin{align}
    P(e;\bm{\hat\phi})-\inf_{\bm\phi} P(e;\bm{\phi}) &= \int_{\lambda\le l(x) \le 1} |P(-1|x)-P(+1|x)|P(x) \, \dif x \nonumber \\
     & = \int_{\lambda\le l(x) \le 1} |1-l(x)|P(-1|x)P(x) \, \dif x \nonumber \\
     & \le (1-\lambda)\int_{\lambda\le l(x) \le 1} P(-1,x) \, \dif x \nonumber \\
     & \le (1-\lambda)\int_x P(-1,x) \, \dif x \nonumber \\
     & = (1-\lambda)P(-1) \nonumber \\
     & = \frac{1-\lambda}{2} \nonumber \\
     & = \frac{\hat\phi_{-1}-\phi^*_{-1}}{2\phi^*_{+1}\hat\phi_{-1}} \nonumber \\
     & \le \frac{1}{2\phi^*_{+1}\phi^*_{-1}} (\hat\phi_{-1}-\phi^*_{-1}) \nonumber \\
     & = \frac{1}{2\phi^*_{+1}\phi^*_{-1}} |\hat\phi - \phi^*|,
\end{align}
where the final inequality is justified by the condition that $\hat\phi_{-1}\ge\phi^*_{-1}$ due to $\lambda\le1$.

\end{proof}

\end{document}